%% file: arxiv.tex
\documentclass{article} 
\usepackage{arxiv,times}
\usepackage{newtxmath}

\input{math_commands.tex}

\usepackage[colorlinks=true,citecolor=blue]{hyperref}
\usepackage{url}

\usepackage{xcolor}         
\usepackage{algorithmic}
\usepackage{algorithm}
\usepackage{booktabs}       
\usepackage{wrapfig}
\usepackage{enumitem}
\usepackage{xspace}

\usepackage{cleveref}
\usepackage{graphicx}
\Crefname{figure}{Figure}{Figures}
\crefname{figure}{Figure}{Figures}
\Crefname{section}{Section}{Sections}
\crefname{section}{Section}{Sections}

\newcommand{\ev}{\mathcal{Z}}          
\newcommand{\lik}{\mathcal{L}}

\title{Improving Gradient-guided Nested Sampling for Posterior Inference}

\author{Pablo Lemos\\
Mila -- Qu\'ebec Artificial Intelligence Institute\\
Universit\'{e} de Montr\'{e}al\\
Ciela Institute \\
CCA -- Flatiron Institute \\
\texttt{pablo.lemos@umontreal.ca}\\
\And
Nikolay Malkin\\
Mila -- Qu\'ebec Artificial Intelligence Institute \\
Universit\'e de Montr\'eal\\
\texttt{nikolay.malkin@mila.quebec}\\
\And
Will Handley \\
Cavendish Laboratory \\
Kavli Institute for Cosmology \\
Gonville and Caius College, University of Cambridge \\
\And 
Yoshua Bengio \\
Mila -- Qu\'ebec Artificial Intelligence Institute \\
Universit\'{e} de Montr\'{e}al\\
Canadian Institute for Advanced Research (CIFAR) \\
\And 
Yashar Hezaveh \\
Mila -- Qu\'ebec Artificial Intelligence Institute\\
Universit\'{e} de Montr\'{e}al\\
Ciela Institute \\
\And 
Laurence Perreault-Levasseur \\ 
Mila -- Qu\'ebec Artificial Intelligence Institute\\
Universit\'{e} de Montr\'{e}al\\
Ciela Institute \\
}

\makeatletter
\def\section{\@startsection {section}{1}{\z@}{-0.3ex}{0.3ex}{\large\sc\raggedright}}
\def\subsection{\@startsection{subsection}{2}{\z@}{-0.2ex}{0.2ex}{\normalsize\sc\raggedright}}
\def\subsubsection{\@startsection{subsubsection}{3}{\z@}{-0.1ex}{0.1ex}{\normalsize\sc\raggedright}}
\def\paragraph{\@startsection{paragraph}{4}{\z@}{0ex}{-1em}{\normalsize\bf}}
\def\subparagraph{\@startsection{subparagraph}{5}{\z@}{0ex}{-1em}{\normalsize\sc}}
\makeatother

\newcommand{\algoname}{\texttt{GGNS}\xspace} 

\iclrfinalcopy 
\begin{document}

\maketitle
\vspace{-0.5cm}
\begin{abstract}
    We present a performant, general-purpose gradient-guided nested sampling algorithm, \algoname, combining the state of the art in differentiable programming, Hamiltonian slice sampling, clustering, mode separation, dynamic nested sampling, and parallelization. This unique combination allows \algoname to scale well with dimensionality and perform competitively on a variety of synthetic and real-world problems.
    We also show the potential of combining nested sampling with generative flow networks to obtain large amounts of high-quality samples from the posterior distribution. This combination leads to faster mode discovery and more accurate estimates of the partition function. 
\end{abstract}

\section{Introduction}

Bayesian parameter estimation and model comparison are key to most scientific disciplines and remain challenging problems, especially in high-dimensional and multimodal settings. While traditionally Markov chain Monte Carlo (MCMC) methods have been used to perform Bayesian inference, differentiable programming has enabled the development of new, more efficient algorithms, such as variational inference~\citep{mackay2003information}, Hamiltonian Monte Carlo \citep{duane1987hybrid, neal2011mcmc}, and Langevin dynamics \citep{besag1994comments, roberts1996exponential, roberts1998optimal}; as well as more recent learning-based methods such as the Path Integral Sampler~\citep{zhang2022path} and generative flow networks~\citep{bengio2021gflownet,lahlou2023theory}. 

From the perspective of differential programming, less attention has been paid in recent years to nested sampling~\citep{10.1214/06-BA127, buchner2021nested, ashton2022nested}, which is a widely used algorithm for Bayesian parameter inference and model comparison. Nested sampling has been used in a range of applications in the natural sciences, from cosmology~\citep{mukherjee2006nested,handley2015cosmo} and astrophysics~\citep{lavie2017helios, gunther2021allesfitter} to particle physics~\citep{yallup2022exploring} and biology~\citep{10.1093/sysbio/syy050}. Furthermore, it provides both samples from the posterior distribution and an estimate of the Bayesian evidence, which can be used for model comparison~\citep{marshall2006bayesian} or to test compatibility between datasets~\citep{handley2019quantifying}. 

The key challenge in implementing a nested sampling algorithm is constructing a method that generates samples drawn from the prior, subject to a hard likelihood constraint. Whilst there is a wide variety of publicly available implementations for doing so~\citep[see][for an exhaustive list]{buchner2021nested, ashton2022nested}, these methods are only capable of scaling to hundreds of dimensions~\citep{scalingfrontier}.

One way to improve the performance of nested sampling algorithms is to use the information about the gradient of the likelihood to propose new points. However, gradient-guided sampling on constrained domains is not straightforward. 
Whilst the materials science and chemistry literature has made extensive use of gradient-guided nested sampling~\citep{2017PhRvE..96d3311B,Partay2021,2015AIPC.1641..121H,10.1063/1.4821761}, these methods are often generally bespoke to their physical problems of interest and are not suitable as general-purpose Bayesian samplers. Previous works, such as~\citet{betancourt2011nested, speagle2020dynesty} have shown  the potential of reflective slice sampling~\citep{neal2003slice}, also known as Galilean Monte Carlo (GMC)~\citep{feroz2013exploring} or Hamiltonian slice sampling (HSS)~\citep{zhang2016towards, pmlr-v48-bloem-reddy16}, for general-purpose sampling{, including the case of non-smooth functions~\citep{mohasel2015reflection}}. However, these approaches have found HSS to be \textit{``substantially less efficient (and in general less reliable) than other gradient-based approaches''}.\footnote{Verbatim from the documentation of the {\tt dynesty} algorithm.} Another alternative is recent work in proximal nested sampling~\citep{cai2022proximal,2023arXiv230700056M}, which uses proximal operators to propose new points. However, this method only works for log-concave likelihoods.

In this work, we combine ideas from across the nested sampling literature and learning-based samplers and create a new gradient-guided nested sampling (\algoname) algorithm. The four key differences with previous work are 1) the use of self-tuning HSS in combination with gradients calculated through differentiable programming to propose new points, 2) incorporating parallel updates using ideas from dynamic nested sampling~\citep{higson2019dynamic,speagle2020dynesty,handley2015polychord} to increase the speed of calculations, 3) A novel termination criterion, and 4) cluster identification to avoid mode collapse.
We show that with these changes in combination, our \algoname algorithm scales to significantly higher-dimensional problems without necessitating a proportional increase in the number of live points with respect to dimensionality. This allows \algoname to perform fast and reliable inference for high-dimensional problems.

We show that the \algoname method presented in this work can be used to perform inference in a wide range of problems and that it can be used to improve the performance of existing nested sampling algorithms. Furthermore, we compare our method to existing algorithms for posterior inference and show that it outperforms them, particularly when dealing with highly multimodal distributions. One of the main advantages of the proposed approach is that it requires little hyperparameter tuning and can be used out-of-the-box in a wide range of problems.

Finally, we show the potential of combining nested sampling with generative flow networks~\citep[GFlowNets, ][]{bengio2021flow,bengio2021gflownet}, which are policy learning algorithms that are trained to generate samples from a target distribution and can flexibly be trained off-policy \citep{malkin2023gflownets}. We show how we can use nested sampling to guide GFlowNet training, leading to faster mode finding and convergence of evidence estimates than with traditional GFlowNets. Conversely, we also show how the amortization achieved by GFlowNets can be used to obtain large amounts of high-quality samples from the posterior distribution. 

\section{Background and Related Work}
\label{sec:ns}

\subsection{Nested Sampling}
\label{sec:ns_summary}

Nested sampling is used for estimating the marginal likelihood, also known as the evidence, in Bayesian inference problems~\citep{10.1214/06-BA127}: 
\begin{equation}\label{eq:evidence}
    \mathcal{Z} = \int \mathcal{L}(\theta) \pi(\theta) d\theta,
\end{equation}
where $\mathcal{L}(\theta)$ is the likelihood function, and $\pi(\theta)$ is the prior distribution.
This integral is often intractable due to the high dimensionality and complexity of modern statistical models. In the process of calculating this integral, nested sampling also produces samples from the posterior distribution.

At its core, nested sampling transforms the evidence integral into a one-dimensional nested sequence of likelihood-weighted prior mass, allowing for efficient exploration of the parameter space. The key idea is to enclose the region of high likelihood within a series of nested iso-likelihood contours. This is achieved by introducing a set of {\it live points} distributed within the prior space and successively updating this set by iteratively replacing the point with the lowest likelihood with a new point drawn from the prior while ensuring the likelihood remains above a likelihood threshold.

As nested sampling progresses, it adaptively refines the prior volume containing higher-likelihood regions. By constructing a sequence of increasing likelihood thresholds, nested sampling naturally focuses on the most informative regions of parameter space. Consequently, nested sampling offers several advantages, including robustness to multimodality in posterior distributions, convergence guarantees, and the ability to estimate posterior probabilities and model comparison metrics.  A more detailed review of the algorithm can be found in~\cref{app:review}.

The number of nested sampling likelihood evaluations scales as~\citep{10.1214/06-BA127,scalingfrontier}:
\begin{equation}
   n_{\rm like} \propto n_{\rm live} \times f_{\rm sampler} \times \mathcal{D}_{\rm KL} (\mathcal{P} | \Pi),  
   \label{eq:ns_scaling}
\end{equation}
where $n_{\rm live}$ is the number of live points, $f_{\rm sampler}$ is the efficiency of the live point generation method (the average number of likelihood evaluations required to generate each new sample), and $\mathcal{D}_{\rm KL} (\mathcal{P} | \Pi)$ is the Kullback-Leibler divergence between the posterior and the prior. 

To understand the scaling of nested sampling with dimensionality, we should consider the three terms separately. Here, $\mathcal{D}_{\rm KL} (\mathcal{P} | \Pi)$ is fixed by the problem at hand (so cannot be modified without substantial adjustment of the meta-algorithm~\citep{2022arXiv221201760P}), and is usually assumed to scale linearly with the number of dimensions. $n_\mathrm{live}$ for most algorithms scales linearly with dimensionality for two independent reasons: First since the uncertainty in the log-evidence estimation is approximately~\citep{10.1214/06-BA127}
\begin{equation}
    \sigma(\log Z) \approx \sqrt{{\mathcal{D}_{\rm KL} (\mathcal{P} | \Pi)}/{n_\mathrm{live}}},
    \label{eqn:logZ_error}
\end{equation}
if we wish to keep this constant we must scale $n_\mathrm{live}$ with $\mathcal{D}_{\rm KL} (\mathcal{P} | \Pi)$, which as discussed before scales linearly with dimension. Second, most practical live point generation methods require a minimum number of points to tune their internal parameters (such as ellipsoidal/cholesky decompositions or neural network training), and this minimum number scales with dimensionality. In the next section, we describe $f_\mathrm{sampler}$ scaling.

\subsection{Previous Work}
\label{sec:previous_work}

The key difficulty in nested sampling is that to generate a new point, one needs to sample points from the prior subject to a hard likelihood constraint:
\begin{equation}
    \{\theta\sim\pi : \mathcal{L}(\theta) > \mathcal{L}_*\} \, .
    \label{eqn:likelihood_constraint}
\end{equation}
Broadly, the mechanisms for achieving this fall into two classes: \emph{region sampling} and \emph{step sampling}~\citep{ashton2022nested}. Region samplers have excellent performance in low dimensions, but have a computational cost that scales exponentially with dimensionality 
$f_\mathrm{sampler}\sim\mathcal{O}(e^{d/d_0})$, where $d_0\sim \mathcal{O}(10)$ is both method and problem dependent. 
Step samplers have a live point generation cost that scales linearly with dimensionality $f_\mathrm{sampler}\sim\mathcal{O}(d)$, so are less efficient in low dimensions.

\emph{Region samplers} use the current set of live points to define a proxy that encapsulates the likelihood-constrained region~\cref{eqn:likelihood_constraint}, and then appropriately samples from this proxy. For example \texttt{MultiNest}~\citep{2008MNRAS.384..449F,feroz2009multinest,2019OJAp....2E..10F} achieves this with an ellipsoidal decompsition fit to the current set of live points, \texttt{nessai}~\citep{2021PhRvD.103j3006W,2023MLS&T...4c5011W} trains a normalising flow and \texttt{ultranest}~\citep{2021JOSS....6.3001B} places ellipsoidal kernels on each live point. 

\emph{Step samplers} run a Markov chain starting from one of the current live points, terminating when one has decorrelated from the initial point and then using the final point of the chain as new point. Whilst Skilling~\citep{10.1214/06-BA127} originally envisaged a Metropolis Hastings step, in practice on its own this is a poor choice for sampling from hard-bounded regions. \texttt{proxnest}~\citep{cai2022proximal,2023arXiv230700056M} uses \texttt{prox}-guided Langevin diffusion, \texttt{DNest}~\citep{2016arXiv160603757B} offers a flexible framework for programming one's own stepper, \texttt{neuralnest}~\citep{2020MNRAS.496..328M} uses normalizing flow guided Metropolis steps and \texttt{PolyChord}~\citep{handley2015cosmo,handley2015polychord} uses slice sampling. Finally, \texttt{dynesty}~\citep{speagle2020dynesty} and \texttt{ultranest}~\citep{2021JOSS....6.3001B} offer Python re-implementations of many of the above within a single package, with a default dimensionality-dependent switching between region and path sampling.

Dynamic nested sampling~\citep{higson2019dynamic,speagle2020dynesty} is a variant of nested sampling which proposes eliminating and replacing multiple points at each iteration. It was initially implemented in the {\tt dyPolyChord}\footnote{\url{https://dypolychord.readthedocs.io/en/latest/}} and {\tt dynesty}\footnote{\url{https://dynesty.readthedocs.io/en/stable/}} packages, but now is common to many implementations~\citep{ashton2022nested}. It has two main use-cases; increasing the number of posterior samples generated by nested sampling, and implementing parallelization schemes.

\subsection{Hamiltonian Slice Sampling}\label{sec:hss}

HSS was first introduced in the context of slice sampling~\citep{neal2003slice}, as a variant of Hamiltonian Monte Carlo. As in slice sampling, the algorithm initially selects an {\it initial point} from the current set of live points and a direction. An initial {\it momentum} variable $p_{\rm ini}$, which is a $d$-dimensional array (where $d$ is the dimension of the space), is also defined, typically by randomly sampling a unit vector. The algorithm then proceeds by simulating the \textit{trajectory} of a particle located at the initial point with the chosen initial velocity integrated with some time step $\Delta t$, such that at each step the position of the particle is updated according to $x'=x+p\Delta t $. When the particle goes beyond the slice, it is {\it reflected} back into the slice. This reflection is performed by updating the momentum from $p$ to $p'$ using the equation
{
\begin{equation}\label{eq:reflection}
    p' = p - 2(p\cdot n)n,\quad n:=\nabla\gL(\theta) / \|\nabla\gL(\theta)\|\, ,
\end{equation}
}

where $n$ is the unit vector in the direction of the likelihood gradient and thus the normal vector to an iso-likelihood surface. Note that, because we are only using the direction of the gradient, one can equivalently use the gradient of the log-likelihood, i.e. the score, which is more efficient to compute. We summarize the algorithm in~\cref{app:hss}.

As highlighted in~\citet{neal2003slice}, \cref{eq:reflection} is only exact when the point where the reflection of the trajectory of the particle takes place is exactly on the boundary $\mathcal{L}(\theta) = \mathcal{L}_*$. In practice, we can either use a small tolerance $\epsilon$ to define a neighborhood  around the slice and reflect a trajectory whenever the particle is within this neighborhood, or reflect a trajectory whenever the particle lands at a point outside the boundary. The latter method has a theoretical risk of a particle getting ``stuck'' behind the boundary (in which case the trajectory would be rejected, and a new initial momentum would be chosen). 

HSS (or GMC) has been used for nested sampling before~\citep{betancourt2011nested,feroz2013exploring,speagle2020dynesty}. However, the {\tt dynesty} implementation and defaults of HSS lacks the efficiency and {reliability} of other sampling methods.
In addition, in these public implementations, a score has to be manually provided since the package is not compatible with modern differentiable programming frameworks.

\section{Contributions}\label{sec:implementation}
In this section, we outline the key combination of ingredients in \algoname we use to significantly improve its performance in high-dimensional settings in comparison with existing publicly available tools. 

In brief: we introduce trimming \& adaptive step size techniques to remove the hyperparameter tuning difficulties that have beset previous implementations, bring in the current state-of-the-art in parallelization and cluster recognition, and implement in differentiable programming which removes the requirement of providing a score function. With these innovations we find that one only needs $\sim\mathcal{O}(1)$ bounces to have decorrelated the chain from the start point, allowing sublinear $f_\mathrm{sampler}$ scaling. Finally, for maximum posterior scaling, the fact that gradients guide the path means one no longer requires $n_\mathrm{live}\sim\mathcal{O}(d)$, giving an in-principle linear scaling with dimensionality for the purposes of posterior estimation.

{This linear scaling has a theoretical basis. For methods such as slice sampling, taking $n$ steps in a $d$ dimensional space leads to sampling an $n$-dimensional subspace. Therefore, we need to reach $\mathcal{O}(d)$ steps to explore the full space. For Hamiltonian slice sampling, on the other hand, every time we use the gradient for a reflection, we get information about the full $d$ dimensional space. Therefore, each step is exploring the full volume, leading to the requirement of $\mathcal{O}(1)$ reflections. This is a similar argument to the better scaling with the dimensionality of Hamiltonian Monte Carlo methods, compared to methods such as random walk Metropolis-Hastings.}

In detail, our contributions are the following. {We include a complete algorithm in Appendix~\ref{app:algorithm} ablation studies showing the importance of various components in Appendix~\ref{app:ablation}.}

\paragraph{Adaptive Time Step Control} We add an adaptive time-step control mechanism in the HSS algorithm. In HSS, particles move in straight lines and eventually reflect off the hard likelihood boundary. To ensure the trajectories between reflections strike a balance between efficiency and accuracy, we introduce the concept of a variable time step, denoted as ${\rm d} t$. This time step is adjusted dynamically during the course of the algorithm. By monitoring the number of reflections, we increase or decrease ${\rm d} t$ to optimize the computational efficiency while maintaining trajectory integrity. This approach, inspired by Neal's work in~\citep{neal2003slice}, enables us to employ larger time steps, thereby reducing the number of reflections without compromising trajectory quality.

\paragraph{Trajectory Preservation} In our second enhancement, we introduce a novel approach to preserving and utilizing trajectory\footnote{{The term trajectory here refers to the states of the chain, not the intermediate states of a Hamiltonian trajectory, as in~\citet{nishimura2020recycling}.}} information during the HSS updates. Specifically, we store all points along the trajectory after a designated number of reflections, where $\mathtt{min\_ref} < \mathtt{max\_ref}$. This archive of trajectory points allows us to efficiently select a new live point {by uniformly sampling} the stored trajectories in a fully parallel manner. We also perturb trajectories by adding some noise $\mathtt{delta\_p}$, to achieve faster decorrelation of the samples.

\paragraph{Pruning Mechanism} To further enhance efficiency, we introduce a ``pruning'' mechanism during the HSS process. Points that have remained outside the slice for an extended duration are identified and removed from consideration. These pruned points are then reset to their initial positions, and new momenta are randomly assigned. This mechanism significantly improves the computational efficiency of the proposed method, {as we do not waste computational resources evaluating the likelihood of points that have drifted far away from the slice.} 

\paragraph{Parallel Evolution of Live Points}
As in \cite{burkoff2012exploring,henderson2014parallelized,martiniani2014superposition,handley2015polychord}, we implement a dynamic approach to live point management, whereby half of the live points are ``killed'' at each iteration and replaced with new points. The new set of live points evolves with our HSS algorithm entirely in parallel, given that the HSS algorithm boils down to simulating simple dynamics for all the live points. This parallelism dramatically accelerates the algorithm's execution.

\paragraph{Mode Collapse Mitigation}
To address the issue of mode collapse, we incorporate a clustering recognition and evolution algorithm as developed and implemented in {\tt PolyChord}~\citep{handley2015polychord}. During the execution of the nested sampling process, we identify clusters of points and keep track of the volume of each cluster. Then, we spawn points proportionally to this volume. This addition helps maintain diversity among live points, preventing them from converging prematurely to a single mode. 

\paragraph{Robust Termination Criterion}
Our final contribution involves the introduction of an alternative termination criterion, which we find to more robust. Unlike previous implementations of nested sampling that rely on the remaining prior volume $\mathcal{X}$, we utilize the property that the quantity $X \mathcal{L}(\theta)$ follows a characteristic trajectory—initially increasing, reaching a peak, and then decreasing. We terminate the algorithm when $X \mathcal{L}(\theta)$ has decreased by a predetermined fraction from its maximum value. {This termination is further explained in~\cref{app:termination}}. This criterion proves to be more resilient to variations in hyperparameters, including the number of live points.

\paragraph{Differentiable Programming}
Whilst nested sampling algorithms written in differential programming languages exist in \texttt{jax}~\citep{2020arXiv201215286A} and \texttt{torch}~\citep{paszke2019pytorch, 2023arXiv230808597A}, these do not make use of gradients in guiding the choice of a new live point,
{Therefore, their choice of using a differentiable programming language is motivated mainly by the advantages of GPU interoperability.}
To our knowledge, ours is the first algorithm utilizing the gradients derived by differentiable programming to guide the choice of a new live point. {Furthermore, this adaptation of nested sampling to hardware intended for modern machine learning workflows, featuring massive parallelization on GPUs; is particularly important in data processing settings that combine nested sampling with deep learning, such as when the prior or likelihood models are given by deep neural networks. We show an example of this when we combine nested sampling and generative flow networks in~\cref{sec:gflownets}.}

We summarize the hyperparameters in~\cref{app:hyperparams} {and provide an ablation study in~\cref{app:ablation}.}

\section{Experiments}

\subsection{Comparison with Other Nested Sampling Methods}\label{sec:comparison}

\begin{figure}[t]
    \centering
    \includegraphics[width=0.99\linewidth]{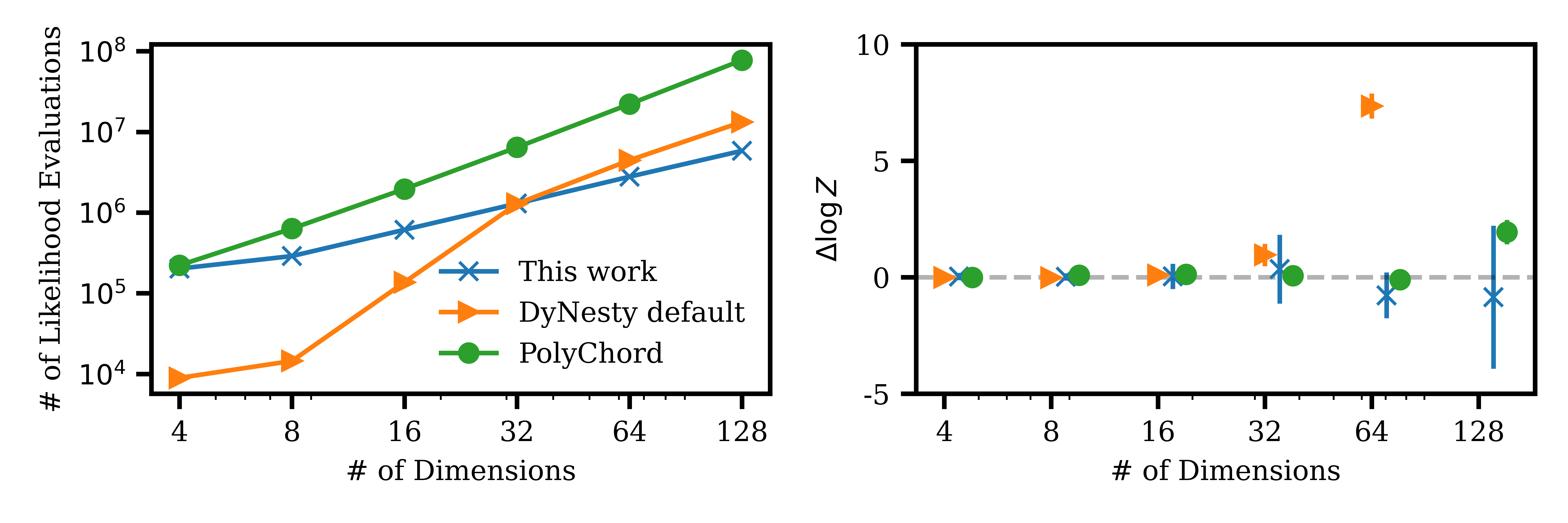}
    \caption{Comparison of likelihood evaluations (left panel) and error in the estimation of $\log \mathcal{Z}$ for different dimensionalities between this work (blue), and other nested sampling algorithms ({\tt PolyChord} in green and {\tt dynesty} in orange; showing more efficient log-log linear scaling while achieving a higher-fidelity estimate. All comparisons are done for a Gaussian likelihood with a diagonal covariance matrix. The error bars show the standard deviation over 10 runs. Error bars for {\tt PolyChord} and {\tt dynesty} are also present, but barely visible. Note that the last point for {\tt dynesty} is not shown, as it is too large to fit in the plot.}
    \label{fig:compare_ns}
\end{figure}

We compare the performance of gradient-guided nested sampling with two popular nested sampling algorithms, already introduced in~\cref{sec:ns}: {\tt PolyChord} and {\tt dynesty}. 
We use the same likelihood function for all algorithms, which is a Gaussian likelihood with a diagonal covariance matrix, and therefore has $\mathcal{D}_{\rm KL} (\mathcal{P} | \Pi)\propto d$. 

For \texttt{PolyChord}, since $f_\mathrm{sampler} \propto n_\mathrm{repeats} = 5d$, from \cref{eq:ns_scaling} we therefore expect $n_\mathrm{like} \propto n_\mathrm{live} \times 5 d^2$. For {\tt dynesty}, its default mode swaps between a region sampler with $n_\mathrm{like}\propto n_\mathrm{live} \times e^{d/d_0} d$ in low dimensions to a slice sampler with $n_\mathrm{repeats}=d$, giving $n_\mathrm{like} \propto n_\mathrm{live} \times d^2$. For \algoname, since $f_\mathrm{sampler} \sim \texttt{max\_ref}\sim \mathcal{O}(1)$, we instead expect $n_\mathrm{like}\propto n_\mathrm{live} \times d$.

For demonstrating the various competing effects discussed in \cref{sec:ns_summary,sec:previous_work,sec:implementation}, we set ${n_{live} =200}$, independent of dimensionality. Note that constant $n_\mathrm{live}$ mode is not usually recommended for these samplers, since as discussed in \cref{sec:ns_summary} we need a minimum number of live points to tune the live point generation hyperparameters. Since \algoname uses gradients to guide the choice of live points, it is not restricted in this way.

The results are shown in~\cref{fig:compare_ns}. We observe the scaling expected from the discussion above. At constant $n_\mathrm{live}$, $\texttt{PolyChord}$ has quadratic scaling with dimensionality, providing good evidence estimates until the dimensionality becomes similar to the $n_\mathrm{live}=200$. $\texttt{dynesty}$ is most efficient but exponentially scaling in low dimensions, and swaps to quadratic scaling in higher dimensions when it moves over to slice sampling, at a lower constant than \texttt{PolyChord} due to its default $n_\mathrm{repeats}=d$ in comparison with $5d$. Note however that this factor of $5$ default efficiency is traded off against poor evidence estimates, even in low dimensions, once it is in slice sampling mode.

\algoname, as predicted, has by far the best (linear) scaling and performs evidence estimation accurately even as the dimensionality approaches the number of live points since its live point generation is guided by gradients rather than the other live points. Note, however, that as expected from \cref{eqn:logZ_error} the error increases with the square root of the dimensionality at fixed $n_\mathrm{live}$.

\subsection{Calculation of Evidence}
\label{sec:evidence}

\begin{table}[t]
\begin{minipage}{0.44\linewidth}
    \caption{Log-evidence function estimation bias (mean and standard deviation over 10 runs). The first rows are from our method, while the rest are from \cite{zhang2022path, lahlou2023theory}. Note that the last three methods are using importance-weighted bound $B_{\rm RW}$. In bold font, we show the estimates that are unbiased at the one standard deviation level.}
\end{minipage}
\hfill
\begin{minipage}{0.54\linewidth}
    \resizebox{1\linewidth}{!}{
    \begin{tabular}{@{}lcc}
    \toprule
    Method&Gaussian mixture&Funnel\\\midrule
    HMC & $-1.876\pm0.527$ & $-0.835\pm0.257$ \\
    SMC & $-0.362\pm0.293$ & $-0.216\pm0.157$ \\
    On-policy PIS-NN & $-1.192\pm0.482$  & ${\bf -0.018\pm0.020}$ \\
    Off-policy GFlowNet TB & ${\bf -0.003\pm0.011}$ & $-0.026\pm0.020$ \\
    On-policy GFlowNet TB & $-1.301\pm0.434$ & ${\bf -0.012\pm0.108}$ \\\midrule
    Ours & ${\bf 0.029\pm0.132}$ & ${\bf -0.051 \pm 0.353}$\\
    \bottomrule
    \end{tabular}
    }
\end{minipage}
    \label{tab:comparison}
\end{table}  

The calculation of the Bayesian evidence is a good way to evaluate the performance of inference algorithms. In this section, we confirm the performance of gradient-guided nested sampling with other methods to sample from a target density. We compare with the following methods: Hamiltonian Monte Carlo~\citep[HMC,][]{mackay2003information,hoffman2014no}, {Sequential Monte Carlo ~\citep[SMC,][]{halton1962sequential, gordon1993novel, chopin2002sequential, del2006sequential}}, Path Integral Sampler~\citep[PIS,][]{zhang2022path} and generative flow networks ~\citep[GFlowNet][]{bengio2021flow,bengio2021gflownet,lahlou2023theory}. {For SMC, the settings follow the code release of~\citet{arbel2021annealed}\footnote{\url{https://github.com/deepmind/annealed_flow_transport.git}.}.} For PIS, we compare with the on-policy version alone, as it obtains better results than the off-policy version. For GFlowNet, we compare with the off and on policy versions, as they perform differently for different tasks, the former being better for multimodal distributions as it is better at exploration, and the latter requiring less samples to converge. We focus on GFlowNets trained with the trajectory balance loss~\citep{malkin2022trajectory}.

We compare these methods with \algoname in two tasks, already introduced in~\citep{hoffman2014no,hoffman2019neutra,zhang2022path,lahlou2023theory}: The first one is the {\bf funnel distribution}, which is a 10D distribution with a funnel shape. The second one is a {\bf Gaussian mixture} in 2-dimension, which consists of a mixture of 9 mode-separated Gaussians. 

As our benchmark, we use the accuracy of the estimate in the log-evidence, or log-partition function. We report the mean and standard deviation of the estimation bias over 10 independent runs in~\cref{tab:comparison}. We observe that gradient-guided nested sampling obtains unbiased estimates in both tasks, something that does not happen for any of the other methods studied in this work. While our standard deviation is higher than that of other methods, these can be reduced by adjusting the hyperparameters of our method. However, \cref{eqn:logZ_error} shows that the nested sampling log-evidence {error can} only be reduced sublinearly by increasing the number of live points $n_\mathrm{live}$, which increases the computational cost. We cannot therefore expect substantial improvements in \algoname log-evidence error bars without innovations in the nested sampling algorithm itself.

\subsection{Image Generation}

\begin{figure}
   \centering
      \begin{minipage}{0.54\textwidth}
   \caption{{\bf First row:} The true image and noise that we aim to reconstruct. {\bf Second row:} The mean from out gradient-guided nested sampling and the standard deviation. We see how the \algoname posterior matches the expected one.
   \label{fig:lensing}}
   \end{minipage}\hfill
\begin{minipage}{0.45\textwidth}
   \includegraphics[width=\linewidth,trim=40 40 280 30,clip]{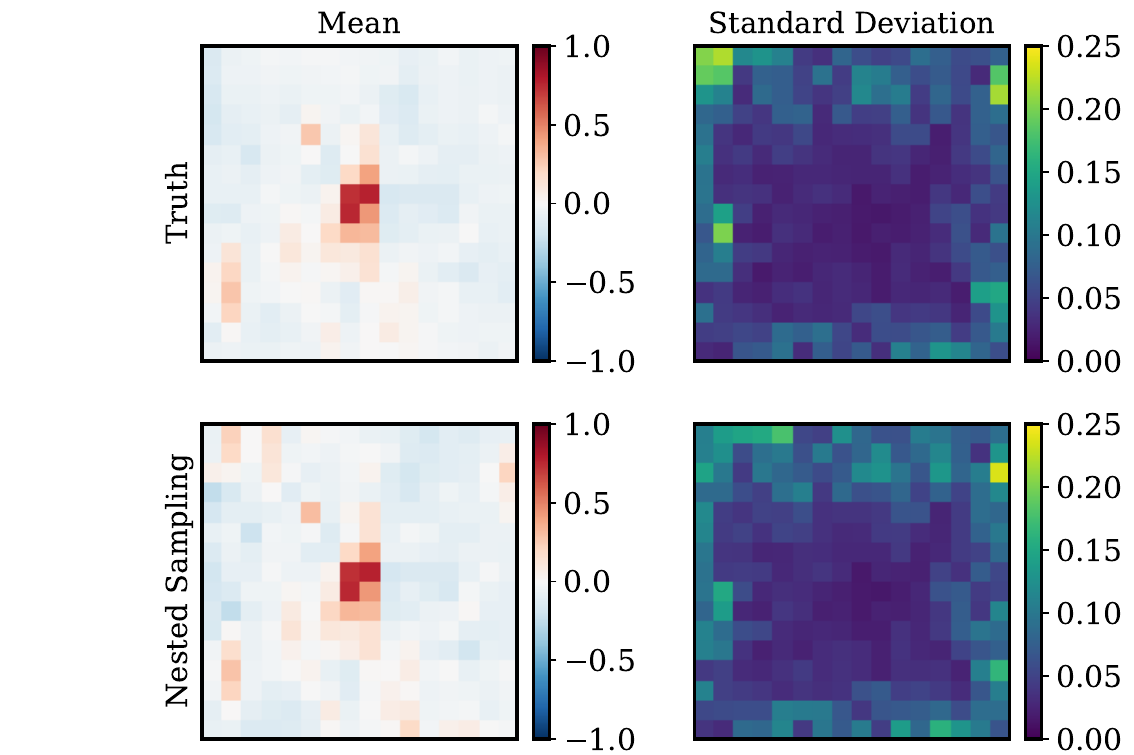}
   \end{minipage}
\end{figure}

We also tested the performance of $\tt{GGNS}$ on a high-dimensional problem, sampling the posterior distribution over image pixels. To do this, we chose the problem of inferring the pixel values of  background galaxies in strong gravitational lensing systems~\cite[e.g.,][]{adam2022posterior}. We assumed a correlated (and non-zero mean) normal prior distribution for the background source. A sample from the prior was generated (representing the background galaxy) and was distorted by a the potential of a foreground lens. Gaussian noise was then added to produce a noisy simulated data. Given the data, the posterior of a model (a pixelated image of the undistorted background source) could be calculated by adding the likelihood and the prior terms. Furthermore since the model is perfectly linear (and known) and the noise and the prior are Gaussian, the posterior is a high-dimensional Gaussian posterior that could be calculated analytically, allowing us to compare the samples drawn with \algoname with the analytic solution.  

\cref{fig:lensing} shows a comparison between the true image, and its noise, and the one recovered by \algoname. We see that we can recover both the correct image, and the noise distribution. We emphasize that this is a uni-modal problem and that the experiment's goal is to demonstrate the capability of \algoname to sample in high dimensions (in this case, $256$), such as images, and to test the agreement between the samples and a baseline analytic solution.

\subsection{Synthetic Molecule Task}

\begin{figure}[t]
\begin{minipage}{0.49\textwidth}
    \caption{The error in the estimate of the normalization of the reward function for the torus task, as a function of the dimensionality of the problem. We see that \algoname obtains consistent estimates of the normalization, even in high dimensional settings.}
\end{minipage}\hfill
\raisebox{-1em}{
\begin{minipage}{0.49\textwidth}
\includegraphics[width=\linewidth]{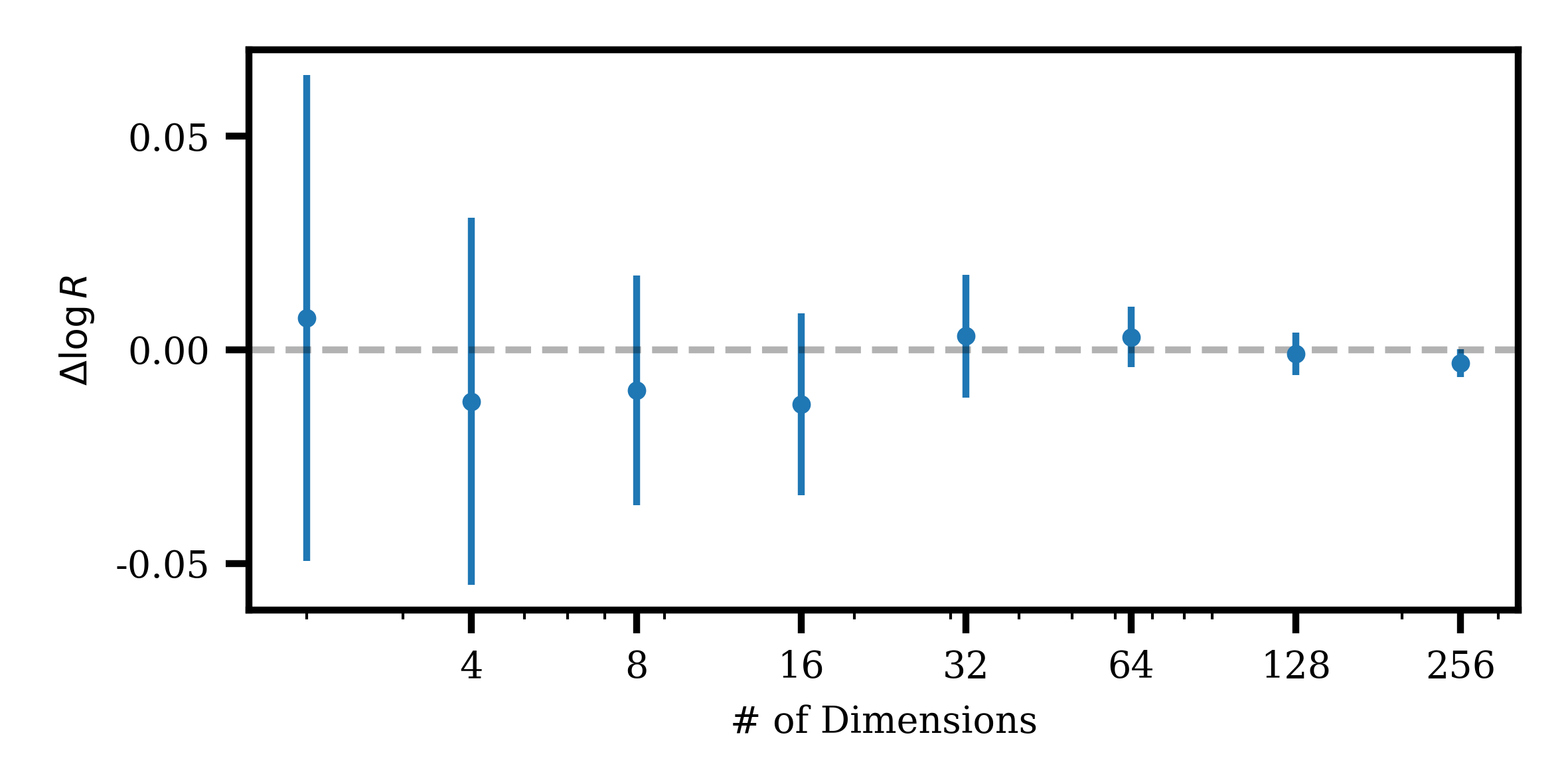}
\end{minipage}
}
    \label{fig:molecules_logr}
 \end{figure}

\begin{figure}[t]
    \centering
    \includegraphics[width=0.49\linewidth]{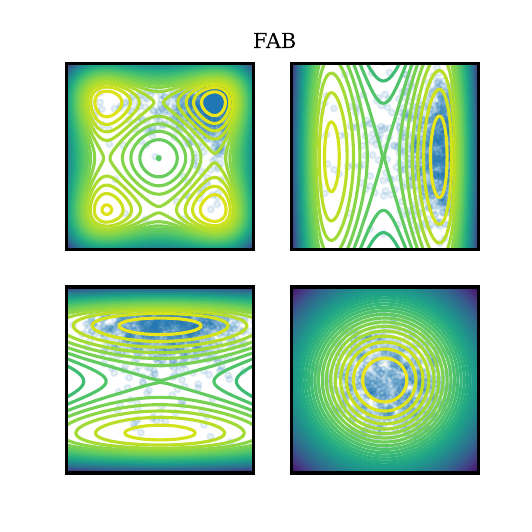}
    \includegraphics[width=0.49\linewidth]{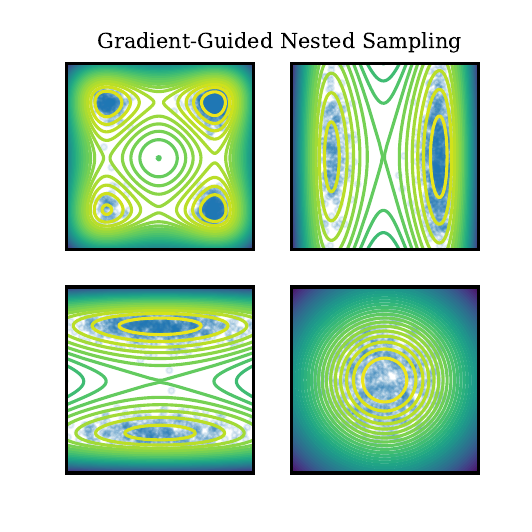}
    \caption{Contour lines for the target distribution, and samples, for the first four dimensions of the 32-dimensional ``Many Wells" problem. On the left, we show results for Flow Annealed Importance Sampling Boostrap (FAB) {with a replay buffer}, and on the right for \algoname. Unlike FAB, \algoname recovers all modes.
    \label{fig:wells_reward}}
 \end{figure}

Finally, we test \algoname on task, inspired by molecular conformations\footnote{To apply this to real molecular configurations, we need a fully differentiable chemical simulator. We leave this for future work}. First, we build a reward function on an $n$ dimensional torus, which extends the reward function introduced in~\cite{lahlou2023theory} to higher dimensional spaces. We define the reward function as:

\begin{equation}
    \label{eq:torus}
    R_n (\mathbf{x}, \alpha, \beta, c) = \left( \sum_{i \ {\rm even}}^n \sin(\alpha x_i) + \sum_{j \ {\rm odd}}^n \cos(\beta x_j) + c \right)^3, \qquad x_i \in [0, 2 \pi). 
\end{equation}
This reward function models the multimodality we expect in molecular conformations, but has the advantage of having a normalization that can be calculated analytically, as detailed in~\cref{app:torus}. This means that we can {assess} the accuracy of \algoname by comparing the estimated normalization with the true value in high dimensional settings. The results are shown in~\cref{fig:molecules_logr}, where we see that \algoname obtains consistent estimates of the normalization, even in high dimensional settings.

We also compare our method to Flow Annealed Importance Sampling Boostrap~\citep[FAB, ][]{midgley2022flow}{, with a replay buffer}. This method has achieved state of the art results in sampling tasks, and was already applied to the Boltzmann distribution of the alanine dipeptide molecule. We use \algoname in two {synthetic} tasks first introduced in~\cite{midgley2022flow}: A mixture of 40 Gaussians in two dimensions, and the 32-dimensional ``Many Well'' problem~\footnote{We use the publicly available implementations of these reward functions  at \href{https://github.com/lollcat/fab-torch}{this URL}.}. The Many Well problem is a particularly challenging one, due to the high its high dimensionality. We show results for the first four dimensions in~\cref{fig:wells_reward}. We see how \algoname does an even better job than FAB at recovering all existing modes. We show visual comparison for the torus reward~\cref{eq:torus} and the mixture of 40 Gaussians in~\cref{app:visual}.

\section{Combination with Generative Flow Networks}\label{sec:gflownets}

We show how the samples obtained from the proposed nested sampling procedure can augment amortized sampling algorithms, such as the generative flow networks considered in \cref{sec:evidence}. In \citet{lahlou2023theory}, it was shown that Euler-Maruyama integration of a stochastic differential equation (SDE) can be viewed as the generative process of a generative flow network. The drift and diffusion terms of the SDE can be trained as the GFlowNet's forward policy to sample from a target distribution given by an unnormalized density. In particular, GFlowNet objectives can be used to learn the reverse to a Brownian bridge between a target distribution and a point, amounting to approximating the reverse to particular kind of diffusion process. The trajectory balance objective -- which directly optimizes for agreement of the forward and reverse path measures -- is equivalent in expected gradient to the objective of the path integral sampler \citep{zhang2022path} when trained using on-policy forward exploration, but can also be trained using off-policy trajectories to accelerate mode discovery, which was found to be beneficial in \citet{lahlou2023theory,malkin2023gflownets}.

\begin{wrapfigure}[39]{r}{0.5\textwidth}
\centering
\vspace*{-1.5em}
\includegraphics[width=0.95\linewidth]{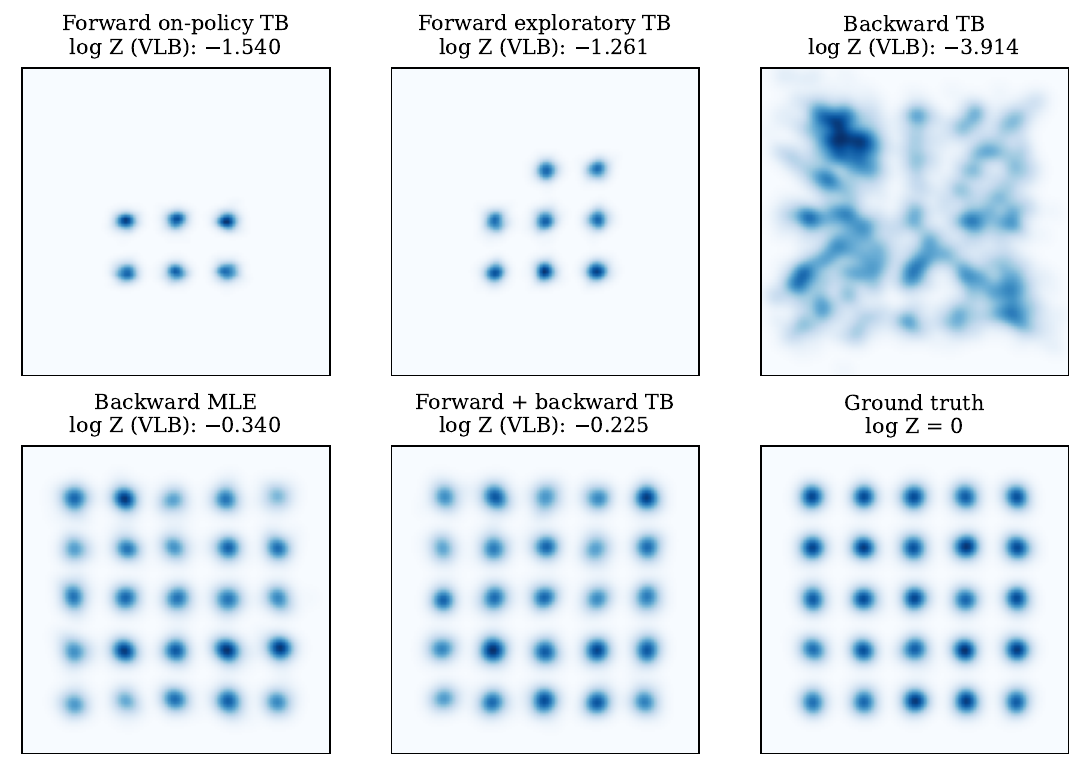}
\includegraphics[width=0.95\linewidth]{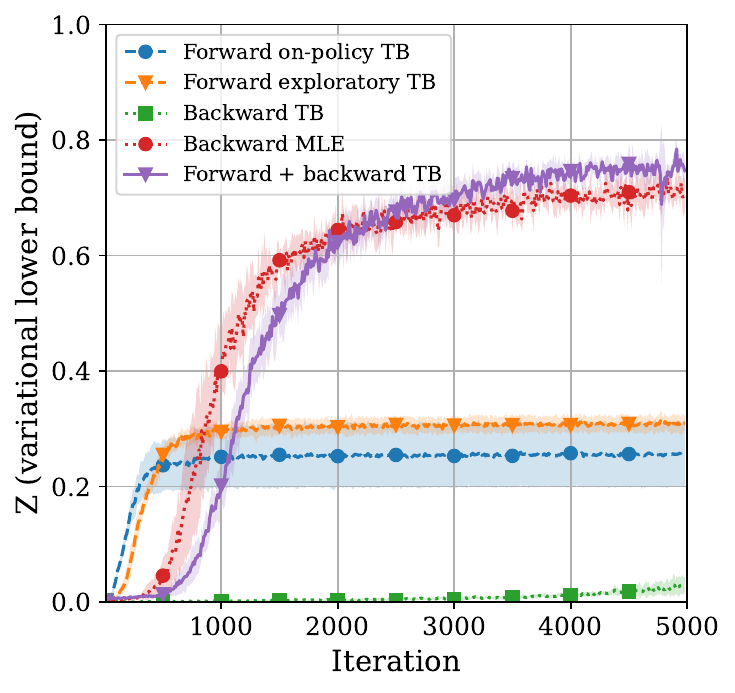}\vspace*{-1.5em}
\caption{\textbf{Above:} KDE plots of samples from trained stochastic control models (ground truth distribution at lower right). Mixing forward sampling with noising trajectories initialized at nested sampling outputs results in all modes being modelled accurately. \textbf{Below:} Variational lower bound on the partition function for the five samplers. The theoretical maximum -- achieved by the Schr\"odinger bridge between the Dirac distribution at the origin and the target distribution -- is $Z=1$.
\label{fig:25gaussians}}
\end{wrapfigure}

Extending the setup of \citet{zhang2022path,lahlou2023theory}, we consider the problem of sampling from a mixture of 25 well-separated Gaussians (see \cref{fig:25gaussians}), with horizontal and vertical spacing of 5 between the component means and each component having variance $0.3$. The learned sampler integrates the SDE $d\mathbf{x}_t=\boldsymbol{\mu}(\mathbf{x},t)\,dt+5\,d\mathbf{w}_t$, where $\boldsymbol{\mu}$ is the output of a neural network (a small MLP) taking $\mathbf{x}$ and $t$ as input, with initial {condition} $\mathbf{x}_0=\boldsymbol{0}$ up to time $t=1$. The reward for $\mathbf{x}_1$ is the density of the target distribution. The neural network architecture and training hyperparameters are the same as in \citet{lahlou2023theory}.

We generate a dataset $\gD$ of 2715 approximate samples from the target distribution first using \algoname, and then we use bootstrapping to generate equally weighted samples, using the bootstrapping algorithm in~\cite{anesthetic}. We consider five algorithms for training the SDE drift $\boldsymbol{\mu}$:
\begin{enumerate}[label=(\arabic*),left=0pt,nosep]
\item \textbf{On-policy TB:} We train $\boldsymbol{\mu}$ by optimizing the trajectory balance objective on trajectories {obtained} by integration of the SDE being trained (equivalent to the path integral sampler objective and to minimization of the KL divergence between forward and reverse path measures).
\item \textbf{Exploratory TB:} We optimize the trajectory balance objective on trajectories obtained from a noised version of the SDE, which adds Gaussian noise with standard deviation $\sigma$ to the drift term at each step. Consistent with \citet{lahlou2023theory}, we linearly reduce $\sigma$ from $0.1$ to 0 over the first 2500 training iterations. Such exploration is expected to aid in discovery of modes.
\item \textbf{Backward TB:} We optimize the trajectory balance objective on trajectories sampled from the reverse (diffusion) process begun at samples from $\gD$.
\item \textbf{Backward MLE:} We sample trajectories from the reverse process begun at samples from $\gD$ and train $\boldsymbol{\mu}$ so as to maximize the log-likelihood of these trajectories under the forward process. This objective amounts to training a diffusion model or score-based generative model \citep{song2019generative,ho2020ddpm} on $\gD$, as the optimal $\boldsymbol{\mu}$ is the score of the target distribution convolved with a Gaussian and appropriately scaled.
\item \textbf{Forward + backward TB:} We optimize the trajectory balance objective both on trajectories obtained by integrating the SDE forward from samples from $\gD$ and on reverse trajectories begun at samples from $\gD$. This method resembles the training policy used by \citet{zhang2022generative}.
\end{enumerate}

KDE plots of samples from the trained models, as well as training metric plots, are shown in \cref{fig:25gaussians}. Training with on-policy TB alone (1) results in mode collapse, a typical effect of training with a {reverse} KL objective \citep{malkin2023gflownets}. We see that while noise introduced in forward exploration (2) helps mode discovery, it is insufficient for all modes to be found. Training using trajectory balance on backward trajectories (3) results in spurious modes, as the model is unlikely to see states that are far from those seen along reverse trajectories from $\gD$ during training. Maximum (4) discovers all modes of the distribution, as they are represented in $\gD$, but closer inspection reveals that they are not modeled as accurately; this effect is more pronounced when the dataset $\gD$ is small. The best sampling performance is reached by models that perform a mix of forward exploration and reverse {trajectories} from the dataset samples.

It is important to note that with well-tuned exploratory policies -- as in (2) -- it is possible to coax the model into discovering all of the modes and modeling them with high fidelity. However, the model is highly sensitive to the exploration parameters: if the exploration rate is too high or not reduced slowly enough, the model is slow to converge and blurs of `fattens' the modes, while an exploration rate that is too low results in mode collapse. On the other hand, mixing forward exploration with backward trajectories from the approximate samples allows the sampler to model all of the modes accurately without such tuning. Notably, we found that the forward trajectories in (5) can be sampled either on-policy or from a tempered policy, with little difference in performance.

\section{Discussion and Conclusions}

We have introduced a new nested sampling algorithm based on Hamiltonian Slice Sampling. Gradient-guided nested sampling improves on previous nested sampling algorithms by removing the linear dependence of the number of live points on dimensionality. It also makes use of the power of differentiable programming frameworks and parallelization for significant speed improvements.
We have shown that the proposed method scales much better with dimensionality than other nested sampling algorithms, thanks to the use of gradient information. This better scaling allows us to apply nested sampling in high-dimensional problems that were too computationally expensive for previous methods. We have also shown that \algoname can be combined with generative flow networks to obtain large numbers of samples from complex posterior distributions. Applications of \algoname to difficult real-world inference problems, both on its own and in combination with amortized sampling methods, are left for future work.

\section*{Reproducibility Statement}

An implementation of \algoname in {\tt PyTorch}~\citep{paszke2019pytorch}, along with notebooks to reproduce the results from the experiments, is available at \url{https://github.com/Pablo-Lemos/GGNS}.

\section*{Acknowledgments}

P.L. would like to thank the Simons Foundation for their generous support.
This research was made possible by a generous donation by Eric and Wendy Schmidt with the recommendation of the Schmidt Futures Foundation.
The work is in part supported by computational resources provided by Calcul Quebec and the Digital Research Alliance of Canada. Y.H. and L.P. acknowledge support from the National Sciences and Engineering Council of Canada grant RGPIN-2020-05102, the Fonds de recherche du Qu{\'e}bec grant 2022-NC-301305 and 300397, and the Canada Research Chairs Program.
Y.B. and N.M. acknowledge funding from CIFAR, Genentech, Samsung, and IBM.

\bibliography{arxiv}
\bibliographystyle{iclr2024_conference}

\appendix
\section{Nested Sampling Review}
\label{app:review}

\begin{algorithm}
    \caption{A simple nested sampling algorithm. Note that more sophisticated implementations include clustering of live points, a calculation of the error in the estimate of $\log Z$, and other techniques to improve performance.}
    \label{alg:ns}
    \begin{algorithmic}[1]
        \STATE Initialise $n_{\rm live}$ live points from the prior $\pi(\theta)$.
        \STATE Initialise an empty set of dead points.
        \STATE Evaluate the likelihood $\mathcal{L}_i = \mathcal{L}(\theta_i)$ for each live point.
        \STATE Set $X = 1$.
        \STATE Set $Z = 0$.
        \WHILE {$X_{N} > {\rm tol} \cdot Z$}
            \STATE Select the live point with the lowest likelihood $\mathcal{L}_j$, and move it from the set of live points to the set of dead points.
            \STATE Sample a new point $\theta_{\rm new}$ from the prior $\pi(\theta)$, under the condition $\mathcal{L}(\theta_{\rm new}) > \mathcal{L}_j$. 
            \STATE Set $Z \rightarrow Z + \frac{1}{n_{\rm live} + 1} X \mathcal{L}_j$.
            \STATE Set $X \rightarrow X {n_{\rm live} \over n_{\rm live} + 1}$.
            \STATE Add $\theta_{\rm new}$ to the set of live points.
        \ENDWHILE
        \FOR {i = 1, ..., $n_{\rm live}$}
            \STATE Select the live point with the lowest likelihood $\mathcal{L}_j$, and move it from the set of live points to the set of dead points.
            \STATE Set $Z \rightarrow Z + {1 \over n_{\rm live} + 1} X \mathcal{L}_j$.
            \STATE Set $X \rightarrow X {n_{\rm live} \over n_{\rm live} + 1}$.
        \ENDFOR
    \end{algorithmic}
\end{algorithm}

Nested sampling was initially introduced as a method to calculate the Bayesian evidence or marginal likelihood: 

\begin{equation}\label{eq:evidence_app}
    \mathcal{Z} = \int \mathcal{L}(\theta) \pi(\theta) d\theta,
\end{equation}
where $\mathcal{L}(\theta)$ is the likelihood function, and $\pi(\theta)$ is the prior distribution. 

The key idea of nested sampling is to define a new variable called the {\it cumulative prior mass} or the {\it prior volume} as:

\begin{equation}\label{eq:cum_prior_mass}
    X(\theta) = \int_{\mathcal{L}(\theta') > \mathcal{L}(\theta)} \pi(\theta') d\theta',
\end{equation}
which is the fraction of the prior mass that has a likelihood greater than the likelihood of the current point. This variable is bounded between 0 and 1, and can be used to rewrite the evidence as:

\begin{equation}\label{eq:ns}
    \mathcal{Z} = \int_0^1 \mathcal{L}(X) dX,
\end{equation}
which is a one-dimensional integral. Therefore, we can evaluate the likelihoods of a set of points $\{\theta_i\}$ sorted by their likelihood, and use them to estimate the evidence by approximating the integral in \cref{eq:ns} as a sum:

\begin{equation}\label{eq:ns_sum}
    \mathcal{Z} \approx \sum_i \mathcal{L}(X_i) \Delta X_i,
\end{equation}
where $\Delta X_i = X_{i-1} - X_i$ is the difference in prior volume between the $i$-th and the $(i-1)$-th point. This approximation is exact in the limit of an infinite number of points, and can be used to estimate the evidence to arbitrary precision.

The key idea of nested sampling is the following: We start by sampling a set of $n_{\rm live}$  {\it live points} from the prior distribution. We then find the point with the lowest likelihood, and remove it from the set, adding it to the set of {\it dead points}. We then replace it with a new point sampled from the prior, subject to the constraint that its likelihood is greater than the likelihood of the point that was removed. This means that, while it is unfeasible to calculate $X(\theta)$ for each of the new points exactly, we can approximate it by using the fact that, at each iteration, the prior volume is contracted by approximately: 

\begin{equation}\label{eq:delta_x}
    \Delta X_i \approx \frac{n_{\rm live}}{n_{\rm live} + 1}. 
\end{equation}

This process is repeated until the remaining posterior mass is smaller than some fraction of the current estimate of $\mathcal{Z}$. The set of points that we have sampled can then be used to estimate the evidence using \cref{eq:ns_sum}. 

Furthermore, \algoname can be used for parameter inference. To do that, we assign the following importance weight to each point:

\begin{equation}\label{eq:importance_weight}
    p_i = \frac{\mathcal{L}(w_i)}{Z},
\end{equation}
where $w_i$ is the prior volume of the shell that was used to sample the $i$-th point:

\begin{equation}\label{eq:shell_volume}
    w_i = X_{i-1} - X_i.
\end{equation}

An example implementation of a nested sampling algorithm is shown in \cref{alg:ns}.

\section{Hamiltonian Slice Sampling Algorithm}
\label{app:hss}

\begin{algorithm}
    \caption{Hamiltonian or Reflective Slice Sampling}
    \label{alg:rss}
    \begin{algorithmic}[1]
        \STATE Choose a point $\theta$ from the existing set of live points.
        \STATE Choose a direction $d$.
        \STATE Choose an initial momentum {$p_{\rm ini} ~ \sim \mathcal{N}(0, 1)$}.
        \STATE Set $p = p_{\rm ini}$.
        \STATE Set $x = \theta$.
        \STATE Set $t = 0$.
        \WHILE{$t < T$}
            \STATE Set $x = x + p \ {\rm d} t$.
            \IF{$x$ is outside the slice}
                \STATE Take $n = \nabla\gL(\theta) / \|\nabla\gL(\theta)\| $
                \STATE Set $p = p - 2(p\cdot n)n$.
            \ENDIF
            \STATE Set $t = t + {\rm d} t$.
        \ENDWHILE
        \STATE Set $\theta' = x$.
    \end{algorithmic}
\end{algorithm}

We show an example implementation of the Hamiltonian or Reflective Slice Sampling algorithm in \cref{alg:rss}.

\section{Hyperparameters of \algoname}\label{app:hyperparams}

\begin{table}[h]
\centering
\caption{Log-evidence function estimation bias (mean and standard deviation over 10 runs). The first rows are from our method, while the rest are from \cite{zhang2022path, lahlou2023theory}. Note that the last three methods are using importance-weighted bound $B_{\rm RW}$. In bold font, we show the estimates that are unbiased at the one standard deviation level.}

\begin{tabular}{p{1.5cm}p{8cm}p{1.5cm}}
\toprule
Parameter & Default Value & Description \\
\midrule
${\tt n_{\rm live}}$ & Number of live points. A higher number leads to better mode coverage. & $200$\footnote{We vary this number across different experiments. These changes are always clarified in the corresponding sections. } \\
{\tt tol} & Tolerance. The stopping criterion. \algoname terminates when $\mathcal{L}_i X_i / \max(\mathcal{L}_i X_i) < {\rm tol}$. & $0.01$ \\
${\tt min\_ref}$ & The minimum number of reflections. We sample points after they have reflected of the boundary at least ${\tt min\_ref}$. & $1$ \\
${\tt max\_ref}$ & The maximum number of reflections. We stop each HSS iteration after the point has reflected ${\tt max\_ref}$ times off the boundary. & $3$ \\
${\tt delta\_p}$ & The number of noise added to the momentum at each HSS step, to decorrelate samples faster. & $0.05$ \\

\bottomrule
\end{tabular}

\label{tab:hparams}
\end{table}

Table~\cref{tab:hparams}, shows the different hyperparameters of \algoname. This shows the little tuning required for \algoname to perform unbiased sampling.

\section{Comparison with Flow Annealed Importance Sampling Boostrap}
\label{app:visual}

\begin{figure}[!ht]
    \centering
    \includegraphics[width=0.99\linewidth]{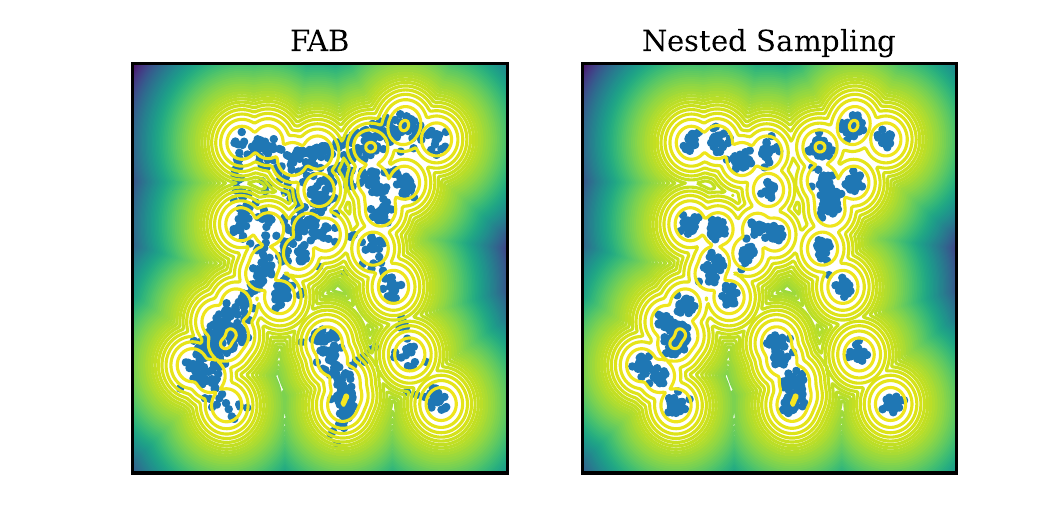}
    \caption{Contour lines for the target distribution, and samples, for the mixture of 40 Gaussians from~\cite{midgley2022flow}. On the left, we show results for Flow Annealed Importance Sampling Boostrap (FAB), and on the right for \algoname.
    \label{fig:fabgaussians}}
\end{figure}

\begin{figure}[t]
\begin{minipage}{0.49\textwidth}
    \caption{Contour lines for the target distribution, and \algoname samples, for the torus reward~\cref{eq:torus}. We do not show Flow Annealed Importance Sampling Boostrap (FAB) samples for this task, as we failed to train it successfully.}
\end{minipage}\hfill
\raisebox{-1em}{
\begin{minipage}{0.49\textwidth}
\includegraphics[width=\linewidth]{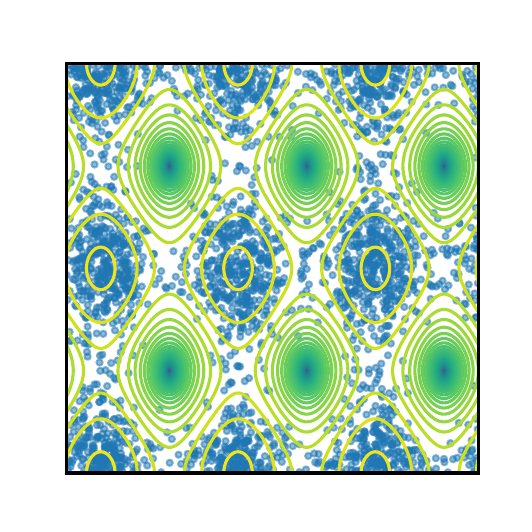}
\end{minipage}
}
    \label{fig:torus}
 \end{figure}

We show the comparison with Flow Annealed Importance Sampling Boostrap (FAB) on the mixture of 40 Gaussians used in~\cite{midgley2022flow}, in~\cref{fig:fabgaussians}. The image shows \algoname samples all the modes of the distribution more accurately than FAB. We also show in~\cref{fig:torus} the results from~\algoname in the torus reward introduced in~\cref{eq:torus}. We see that \algoname can {successfully} sample the distribution. We do not show a comparison with FAB on this task, as we could not easily train it on a torus.

\section{Sampling Complex Distributions}

\begin{figure}[!ht]
    \centering
    \includegraphics[width=0.99\linewidth]{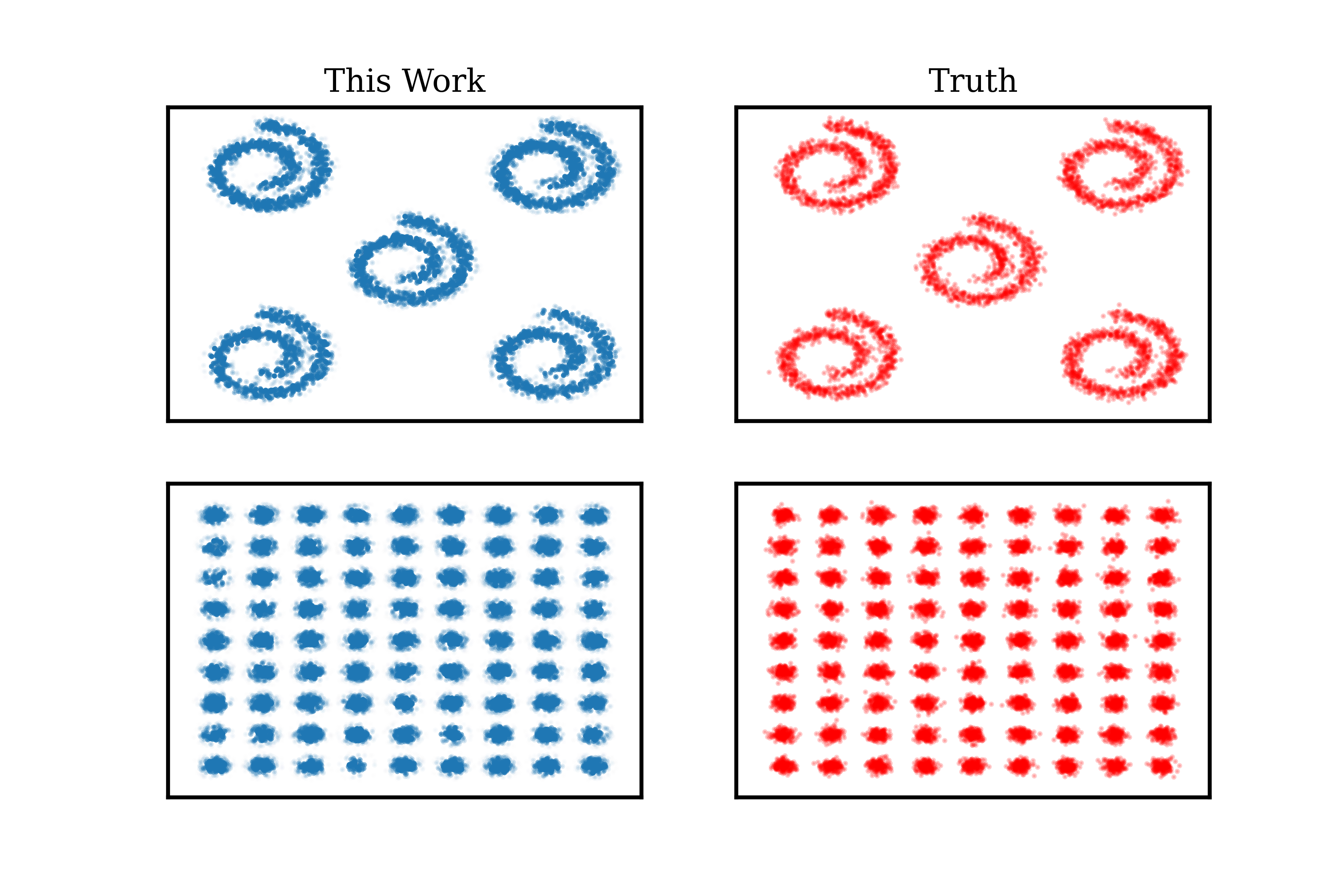}
    \caption{Comparison between the proposed method (left) and the truth (right) on the "five Swiss rolls" distribution. We show our method successfully recovers every model of this highly multi-modal distribution. 
    \label{fig:swissrolls}}
\end{figure}

We further test the capacity of \algoname to model several complex distributions that are for different reasons challenging for inference algorithms. For these examples, we use a visual comparison with samples from the true distribution. We increase the number of live points to $2000$ for these examples, to ensure that we have enough samples to compare with the true distribution. Because nested sampling produces weighted samples, all the figures use an alpha blending of the samples, with the alpha value proportional to the importance weight.

Firstly, we re-use the Gaussian mixture distribution from the previous example, but we increase the number of modes to $81$. This distribution is difficult due to its very high multimodality. The results are shown in the top panel of~\cref{fig:swissrolls}, where we see that our method successfully recovers all modes. 

Secondly, the "five Swiss rolls" example consists of five copies of the "Swiss roll" distribution in two dimensions. It combines multimodality with the difficulty of sampling the complex structure of each mode of the distribution. As shown in the bottom panel of~\cref{fig:swissrolls}, the proposed method successfully samples the distribution.


{
\section{Torus Reward Function}
\label{app:torus}
}

{
\subsection{Normalization}
}

As stated in the main text, for the following reward: 

\begin{equation}
    R_n (\mathbf{x}, \alpha, \beta, c) = \left( \sum_{i \ {\rm even}}^n \sin(\alpha x_i) + \sum_{j \ {\rm odd}}^n \cos(\beta x_j) + c \right)^3, \qquad x_i \in [0, 2 \pi)
\end{equation}
the normalization, defined as: 

\begin{equation}
    Z_n (\alpha, \beta, c) = \int R_n (\alpha, \beta, c) \ \mathrm{d} \mathbf{x}
\end{equation}
This can be found with the following recursive formula: 

\begin{equation}
    Z_n (\alpha, \beta, c) = 2 \pi Z_{n - 1} (\alpha, \beta, c) + 3 \pi c (2 \pi)^{n - 1}, \qquad Z_1 (\alpha, \beta, c) = 2 \pi c^3 + 3 \pi c.
\end{equation}
As long as $\alpha, \beta \in \mathbb{Q}$

The easiest way to prove it is by induction: It is trivially true for $Z_1$:

\begin{equation}
\label{eq:z1}
\int \left( \sin (\alpha x) + c \right)^3 \ \mathrm{d} x = 2 \pi c^3 + 3 \pi c, \qquad \forall \alpha \in \mathbb{Q}, \ \forall c. 
\end{equation}
Note that this also {applies} if we swap a sine for a cosine, which will become relevant later.

So, all that is left is to prove that, if the statement holds for $Z_{n - 1}$, it also holds for $Z_{n}$. Lets assume, without loss of generality, that $n$ is even

\begin{align}
    Z_n (\alpha, \beta, c) &= \int \left( \sum_{i \ {\rm even}}^n \sin(\alpha x_i) + \sum_{j \ {\rm odd}}^n \cos(\beta x_j) + c \right)^3 \ \mathrm{d} x_1 ... \mathrm{d} x_n \\
    &= \int \left[ \int \left( \sin (a x_n) + \sum_{i \ {\rm even}}^{n - 1} \sin(\alpha x_i) + \sum_{j \ {\rm odd}}^{n - 1} \cos(\beta x_j) + c \right)^3 \mathrm{d} x_n \right] \mathrm{d} x_1 ... \mathrm{d} x_{n-1}.
\end{align}
Let's define: 

\begin{equation}
    K \equiv \sum_{i \ {\rm even}}^{n - 1} \sin(\alpha x_i) + \sum_{j \ {\rm odd}}^{n - 1} \cos(\beta x_j) + c.
\end{equation}
Note that $K$ does not depend on $x_n$. With this: 

\begin{align}
    Z_n (\alpha, \beta, c) &= \int \left[ \int \left( \sin (a x_n) + K \right)^3 \mathrm{d} x_n \right] \mathrm{d} x_1 ... \mathrm{d} x_{n-1}.
\end{align}
The key realization here is that the thing inside the bracket is the same as~\cref{eq:z1}. Therefore: 

\begin{align}
    Z_n (\alpha, \beta, c) &= \int \left( 2 \pi K^3 + 3 \pi K \right) \mathrm{d} x_1 ... \mathrm{d} x_{n-1}. \\
    &= 2 \pi \int K^3 \mathrm{d} x_1 ... \mathrm{d} x_{n-1} + 3 \pi \int K  \mathrm{d} x_1 ... \mathrm{d} x_{n-1}
\end{align}

The first integral is simply $Z_{n - 1} (\alpha, \beta, c)$. The second is

\begin{align}
    \int K  \mathrm{d} x_1 ... \mathrm{d} x_{n-1} &= \int \left[ \sum_{i \ {\rm even}}^{n - 1} \sin(\alpha x_i) + \sum_{j \ {\rm odd}}^{n - 1} \cos(\beta x_j) + c \right]  \mathrm{d} x_1 ... \mathrm{d} x_{n-1} \\
    & = (2 \pi)^{n-1} c, 
\end{align}
because all the sine and cosine integrals cancel. Therefore: 

\begin{align}
    Z_n (\alpha, \beta, c) &= 2 \pi \int K^3 \mathrm{d} x_1 ... \mathrm{d} x_{n-1} + 3 \pi \int K  \mathrm{d} x_1 ... \mathrm{d} x_{n-1} \\
    &= 2 \pi Z_{n - 1} (\alpha, \beta, c) + 3 \pi c (2 \pi)^{n - 1}    
\end{align}
Q.E.D.

{
\subsection{Behaviour on different dimensions}
}

\begin{figure}[t]
\begin{minipage}{0.49\textwidth}
    \caption{Change in the Kullback-Leibler divergence with dimensionality, for the molecule example shown in~\cref{fig:molecules_logr}. As described by~\cref{eqn:logZ_error}, the error in the estimate of $\log Z$ is proportional to $\mathcal{D}_{KL} ( \mathcal{P} | \Pi)$, therefore, the unusual decrease of $\mathcal{D}_{KL} ( \mathcal{P} | \Pi)$ with dimensionality, explains the reduced errors on high dimensions for the estimate of $\log Z$.}
\end{minipage}\hfill
\raisebox{-1em}{
\begin{minipage}{0.49\textwidth}
\includegraphics[width=\linewidth]{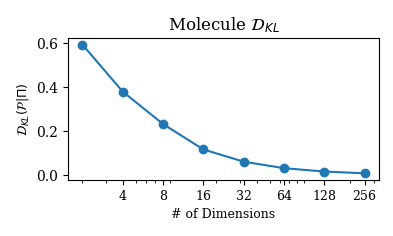}
\end{minipage}
}
    \label{fig:molecule_dfk}
 \end{figure}

{
Our nested sampling $\log Z$ estimates for the synthetic molecule reward, show an unusual pattern of behaviour, as we can see in~\cref{fig:molecules_logr}: The error goes down as the dimensionality increases. This section offers some intuition about why this happens. As described by~\cref{eqn:logZ_error}, the nested sampling error in the estimate of $\log Z$ is proportional to $\sqrt{{\mathcal{D}_{\rm KL} (\mathcal{P} | \Pi)}/{n_\mathrm{live}}}$. Because our problem keeps $n_\mathrm{live}$ fixed, observed behaviour is due to a change in $\mathcal{D}_{\rm KL} (\mathcal{P} | \Pi)$ as the dimensionality increases. 
}


{
$\mathcal{D}_{\rm KL} (\mathcal{P} | \Pi)$ is a measure of how much information we gain when we go from the prior to the posterior. Therefore, in more cases, it increases with dimensionality. However, as shown by~\cref{fig:molecule_dfk}, in this particular example, $\mathcal{D}_{\rm KL} (\mathcal{P} | \Pi)$ goes down with dimensionality, which explains why the error goes down with dimensionality in~\cref{fig:molecules_logr}. The reason why $\mathcal{D}_{\rm KL} (\mathcal{P} | \Pi)$ is likely caused by cancellations in the various terms of the sum~\cref{eq:torus}, as dimensionality increases, but will be further explored in future work. 
}

{\section{Ablation Study} \label{app:ablation}}

{Given the multiple different components included in our algorithm, described in~\cref{sec:implementation}, it is important to perform an ablation study to fully understand how the different parts of the algorithm contribute to the observed improvements in performance. Therefore, this section removes each of the improvements introduced in~\cref{sec:implementation} one by one and analyses the effect of these changes on the sampling performance. We use the task introduced in~\cref{sec:comparison} to perform this comparison unless otherwise specified.}

{
\subsection{Adaptive Time Step Control} 
}
{
\begin{figure}[t]
    \centering
    \includegraphics[width=0.99\linewidth]{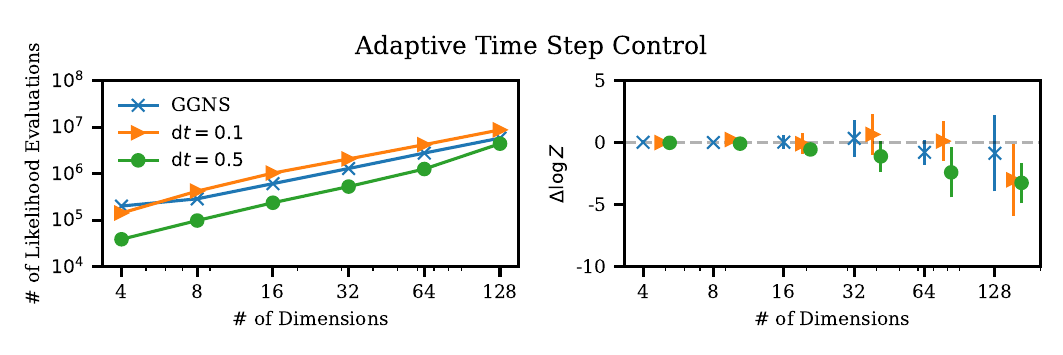}
    \caption{Comparison of likelihood evaluations (left panel) and error in the estimation of $\log \mathcal{Z}$ for different dimensionalities between baseline \algoname (blue), and \algoname without adaptive step control (${\rm d} t = 0.5$ in green and  ${\rm d} t = 0.1$ in orange)}
    \label{fig:ablation_dt}
\end{figure}
}

{
We repeat the analysis of~\cref{sec:comparison} with a fixed time step in the Hamiltonian slice sampling steps instead of using adaptive time step control. We attempt three different time steps: ${\rm d} t = 0.5$ and ${\rm d} t = 0.1$. Note that, in~\cref{sec:comparison}, we start with ${\rm d} t = 0.1$ but adapt it as the algorithm progresses.     
}

{
The results, shown in~\cref{fig:ablation_dt}, show that when we use a large ${\rm d} t$ we can reduce the number of likelihood evaluations, as we achieve the minimum number of reflections faster, but we get a biased estimate of $\log Z$, as we fail to appropriately sample each slice. For small ${\rm d} t$, on the other hand, we get less bias in $\log Z$, but the number of likelihood evaluations goes up. 
}

{
In general, the advantage of the adaptive step is that as the algorithm progresses, the volume of the region defined by~\cref{eqn:likelihood_constraint} we are exploring decreases. Therefore, a step size that is appropriate at a given point will become too large eventually as the algorithm progresses.
}

{
\subsection{Trajectory preservation}
}

{
\begin{figure}[t]
    \centering
    \includegraphics[width=0.99\linewidth]{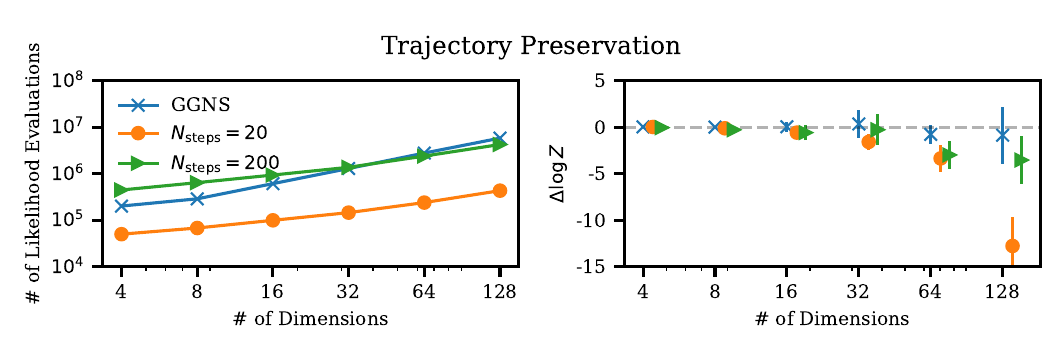}
    \caption{Comparison of likelihood evaluations (left panel) and error in the estimation of $\log \mathcal{Z}$ for different dimensionalities between baseline \algoname (blue), and \algoname without trajectory preservation ($N_{\rm steps} = 200$ in green and  $N_{\rm steps} = 20$ in orange). Both of these lead to correlated estimates of $\log Z$, as shown by the right panel.}
    \label{fig:ablation_nsteps}
\end{figure}
}

{
\begin{figure}[t]
    \centering
    \includegraphics[width=0.99\linewidth]{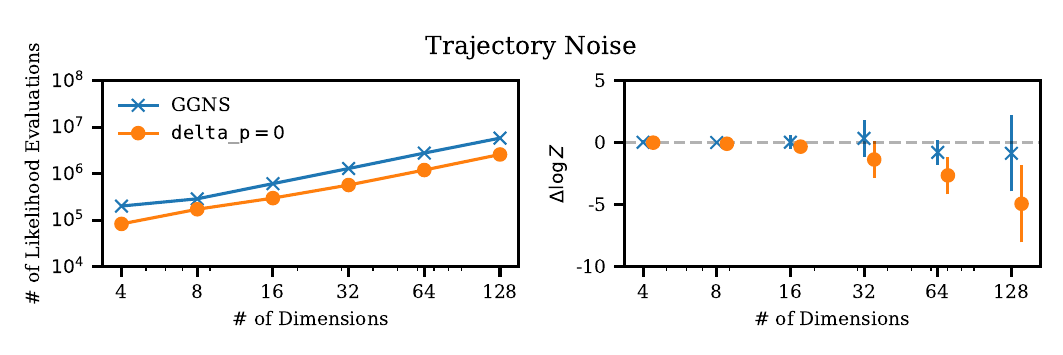}
    \caption{Comparison of likelihood evaluations (left panel) and error in the estimation of $\log \mathcal{Z}$ for different dimensionalities between baseline \algoname (blue), and \algoname without adding noise to the trajectories (orange). Less noise decreases the number of evaluations but leads to a biased $\log Z$ estimate.}
    \label{fig:ablation_noise}
\end{figure}
}

{
\algoname uses a novel approach to sample the trajectories and to ensure that samples are correlated, where we ensure a certain number of boundary reflections. We compare what happens when we use the simpler approach of integrating our trajectory for a fixed number of steps $n_{\rm steps}$ and simply keeping the last sample. We repeat the analysis for $n_{\rm steps} = 20$ and $n_{\rm steps} = 200$. 
}

{
\cref{fig:ablation_nsteps} shows the results. We see that a fixed number of steps leads to a biased estimate of $\log Z$. The argument for this, similarly to what it was for trajectory preservation, is that the volume of the region~\cref{eqn:likelihood_constraint} decreases as the algorithm progresses.
}

{
We also study what happens when we use trajectory preservation but do not add noise ${\tt \delta_p}$ to achieve a faster decorrelation of the samples. The results, shown in~\cref{fig:ablation_noise}, are intuitive: No noise in the trajectories reduces the likelihood evaluations, as trajectories are less noisy but lead to biased evidence estimates, as the samples are not fully decorrelated. 
}

{
\subsection{Mode Collapse Mitigation}
}

{
\begin{figure}[t]
    \centering
    \includegraphics[width=0.99\linewidth]{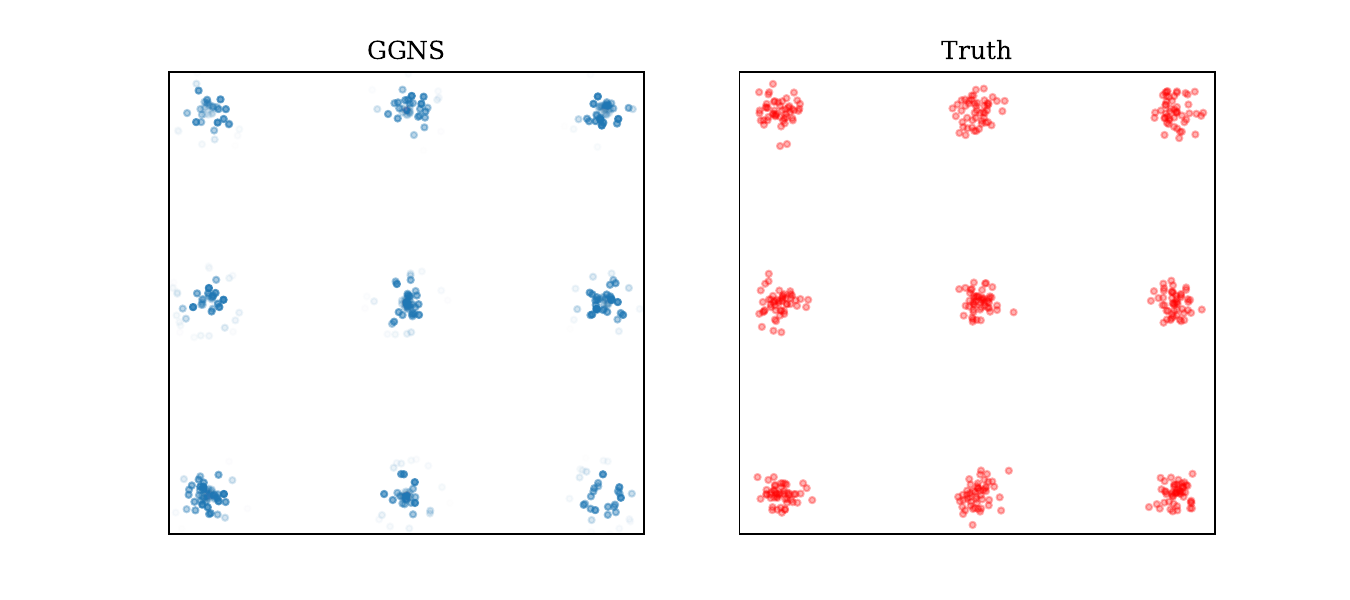}
    \caption{The distribution used for our ablation study on mode collapse mitigation, true samples on the right, and samples with GGNS ($n_{\rm live} = 100$) on the left.}
    \label{fig:9gaussians}
\end{figure}
}

\begin{table}[t]
    \caption{The average number of modes found over $10$ nested sampling runs, sampling the distribution shown in~\cref{fig:9gaussians}, with and without mode collapse mitigation, for varying number of live points.}
    \begin{tabular}{@{}lcccc}
    \toprule
    Method& $n_{\rm live} = 20$ & $n_{\rm live} = 50$ & $n_{\rm live} = 100$ & $n_{\rm live} = 200$\\\midrule
    Without Mode Collapse Mitigation & $3.6$ & $6.4$ & $8.2$ & $8.9$\\
    With Mode Collapse Mitigation & $4.1$ & $6.4$ & $8.4$ & $9$\\
    \bottomrule
    \end{tabular}
    \label{tab:ablation}
\end{table}  

{
To study the effect of this, we need a multimodal distribution. We use a mixture of nine Gaussians, shown in~\cref{fig:9gaussians}. To study the effect of our mode collapse mitigation, we run nested sampling on this problem, with and without this setting. The main hyperparameter that affects the number of modes found, is the number of live points $n_{\rm live}$. Therefore, we run nested sampling on this problem for different values of this hyperparameter. For each configuration, we run the algorithm $10$ times, and count the number of modes found. We define a mode as being found, if at least one of the samples is within a distance $\sigma$ of the center of the mode. 
}

{
The results are shown in~\cref{tab:ablation}. We see how, generally, mode collapse mitigation helps us find a higher number of modes. Although the number of live points is the most important hyperparameter when it comes to mode finding, the ability to find all modes for a fixed $n_{\rm live}$ is higher when using mode collapse mitigation.
}

{
\subsection{Termination Criterion} 
}

{
\begin{figure}[t]
    \centering
    \includegraphics[width=0.99\linewidth]{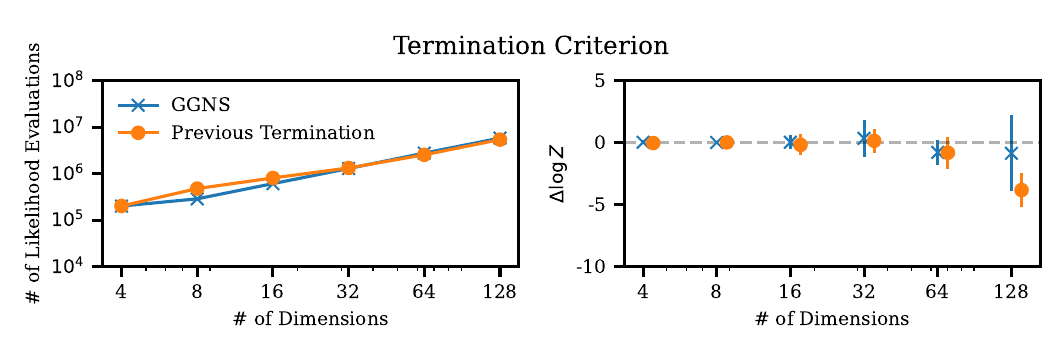}
    \caption{Comparison of likelihood evaluations (left panel) and error in the estimation of $\log \mathcal{Z}$ for different dimensionalities between baseline \algoname (blue), and \algoname using the termination criterion used by {\tt MultiNest}, {\tt PolyChord} and {\tt DyNesty}.}
    \label{fig:ablation_termination}
\end{figure}
}

{
Finally, we repeat the analysis of~\cref{sec:comparison} using the termination criterion used by other nested sampling algorithms such as {\tt DyNesty} and {\tt PolyChord}, in which we terminate the algortihm when $\mathcal{L}_{\rm max} X_i < {\rm tol}$, for some tolerance hyperparameter. We used a tolerance $0.01$, a value often used by nested sampling practitioners.
}

{
We show the results in ~\cref{fig:ablation_termination}. The number of likelihood evaluations appears similar, but at high dimensions, the previous termination leads to a biased estimate of $\log Z$. Indeed, while the number of likelihood estimations is of the same order of magnitude, the termination used by \algoname leads to slightly more evaluations ($ \sim 5.8 \cdot 10^6$ )than the previous one ($ \sim 5.4 \cdot 10^6$ ) for $d = 128$. These $400,000$ evaluations are likely to drive the underestimation of the evidence by the previous method. 
}

{
\subsection{Other changes}
We could not study other changes, such as the pruning mechanism or parallel Evolution of live points, as this would have led to a full rewrite of the algorithm. We leave this study for future work.
}

{
\subsection{Conclusions}
}

{
The main conclusion of this ablation study is that naive changes to \algoname quickly lead to biased sampling. It is the combination of the contributions introduced in~\cref{sec:implementation} that leads to robust evidence on high dimensions. Of course, it is true that for each of these parts, there are settings that will work for any problem, i.e. we can always make the step size small enough, the number of steps high enough, etc. However, the main advantage of our algorithm is that it works without the need for fine-tuning all these parameters.
}

{
\section{\algoname algorithm}\label{app:algorithm}
}

\begin{algorithm}
    \caption{The \algoname algorithm}
    \label{alg:ggns}
    \begin{algorithmic}[1]
        \STATE Initialise $n_{\rm live}$ live points from the prior $\pi(\theta)$.
        \STATE Initialise an empty set of dead points.
        \STATE Evaluate the likelihood $\mathcal{L}_i = \mathcal{L}(\theta_i)$ for each live point.
        \STATE Initiate summary statistics $ \left\{ \overline{Z}, \overline{X}, \overline{Z_p}, ... \right\} $, using~\cref{eq:ini0} to~\cref{eq:ini11}.
        \STATE Set $\Delta (X \mathcal{L})  = 0$.
        \STATE Set $(X \mathcal{L})_{\rm max}  = 0$.
        \STATE Set $n_{\rm clusters} = 1$.
        \STATE Set $dt = dt_{\rm ini}$.
        \WHILE {$\Delta (X \mathcal{L}) > {\rm tol} $} 
            \FOR {i = 1, ..., $n_{\rm clusters}$}
                \STATE Use cluster finding~\cref{alg:ggns_clusters}
                \STATE If new clusters are found, initiate them by splitting the cluster $i$, using~\cref{eq:new0} to~\cref{eq:new6}
            \ENDFOR
            \FOR {j = 1, ..., $n_{\rm live} // 2$}
                \STATE Select the point with the lowest likelihood $\mathcal{L}_j$ and remove them from the set of live points to the set of dead points. 
                \STATE Update the summary statistics, using~\cref{eq:update0} to~\cref{eq:update9}.
                \STATE If a cluster has no points, remove it, and set $n_{\rm clusters} -= 1$
            \ENDFOR
            \STATE Generate the cluster labels for the next points $n_{\rm live} // 2$, proportionally to $\overline{X_p}$
            \STATE Sample $x \sim \theta_{\rm live}$, from the appropriate clusters
            \STATE Use~\cref{alg:ggns_hss}, to get $\theta_{\rm new}$, and ${\tt out\_frac}$, under the condition $\mathcal{L}(\theta_{\rm new}) > \mathcal{L}_j \ \forall \theta_{\rm new}$.
            \IF{${\tt out\_frac} > 0.15$}
            \STATE    Set $dt = dt * 0.9$
            \ELSIF{${\tt out\_frac} < 0.05$}
            \STATE   Set $dt = dt * 1.1$
            \ENDIF
            \STATE Add $\theta_{\rm new}$ to the set of live points.
            \STATE Set $(X \mathcal{L})_{\rm max} = \max \left(X \mathcal{L_{\rm max}}, (X \mathcal{L})_{\rm max} \right)$, where $\mathcal{L}_{\rm max}$ is the maximum likelihood amongst the live points.
            \STATE Set $(\Delta X \mathcal{L}) = X \mathcal{L}_{\rm max} / (X \mathcal{L})_{\rm max}$
        \ENDWHILE
        \FOR {i = 1, ..., $n_{\rm live}$}
            \STATE Select the live point with the lowest likelihood $\mathcal{L}_j$ and move it from the set of live points to the set of dead points.
            \STATE Set $Z = Z + {1 \over n_{\rm live} + 1} X \mathcal{L}_j$.
            \STATE Set $X = X {n_{\rm live} \over n_{\rm live} + 1}$.
        \ENDFOR
    \end{algorithmic}
\end{algorithm}

\begin{algorithm}
    \caption{The cluster finding algorithm used in~\cref{alg:ggns}, for a cluster containing $n_{points}$ points. }
    \label{alg:ggns_clusters}
    \begin{algorithmic}[1]
        \STATE Initialise ${\tt prev\_sizes} = {\tt None}$.
        \FOR {k = 2, ... , $n_{points}$ }
            \STATE Run k-nearest-neighbours (KNN) on the cluster points, with value $k$
            \STATE Set ${\tt cluster\_sizes}$ as the number of points in each KNN cluster
            \IF{${\tt cluster\_sizes} = {\tt prev\_sizes}$}
                \STATE Break
            \ELSE
                \STATE Set ${\tt prev\_sizes} = {\tt cluster\_sizes}$
            \ENDIF
        \ENDFOR
        \RETURN The number of KNN clusters. 
    \end{algorithmic}
\end{algorithm}

\begin{algorithm}
    \caption{The Hamiltonian slice sampling algorithm used in~\cref{alg:ggns}, starting from $n$ points with position $x$, and with step size $dt$; and with a likelihood barrier $\mathcal{L}_{\rm min}$}
    \label{alg:ggns_hss} 
    \begin{algorithmic}[1]
        \STATE Set ${\tt num\_out\_steps} = 0$, ${\tt num\_in\_steps} = 0$
        \STATE Set $p ~ \sim \mathcal{N}(0, 1)$.
        \STATE Set ${\tt num\_reflections}[1, ...,  n] <- 0$
        \STATE Set  ${\tt x\_saved} = \{ \}$
        \WHILE{$\min({\tt num\_reflections}) < {\tt max\_reflections}$}
            \STATE Set $x += p * dt$
            \STATE Call the likelihood function, to get $\mathcal{L}$ and $\nabla \mathcal{L}$
            \STATE Set ${\tt outside}[1, ...,  n] = \mathcal{L} < \mathcal{L}_{\rm min}$
            \STATE Take $n = \nabla\gL / \|\nabla\gL)\| $
            \STATE Set $p[{\tt outside}] = p[{\tt outside}] - 2(p\cdot n)n [{\tt outside}]$.
            \STATE Set $\epsilon \sim \mathcal{N}(0, 1)$
            \STATE Set $p = p * (1 + \epsilon * {\tt delta\_p})$
            \STATE Set ${\tt num\_reflections} += {\tt outside}$
            \IF{$\min({\tt num\_reflections}) < {\tt min\_reflections}$}
                \STATE Add $x[\sim {\tt outside}]$ to ${\tt x\_saved}$
            \ENDIF
            \STATE Set ${\tt num\_out\_steps} += \sum({\tt outside})$
            \STATE Set ${\tt num\_in\_steps} += \sum({\tt \sim outside})$
        \ENDWHILE
        \STATE Set ${\tt out\_frac} = {\tt num\_out\_steps} / ({\tt num\_out\_steps} + {\tt num\_in\_steps})$
        \STATE Samples $\theta \sim {\tt x\_saved}$
        \RETURN $\theta$, ${\tt out\_frac}$
    \end{algorithmic}
\end{algorithm}

{
We show the full \algoname algorithm in~\cref{alg:ggns}. Our cluster-finding algorithm and our Hamiltonian slice sampling algorithm are shown in~\cref{alg:ggns_clusters} and~\cref{alg:ggns_hss} respectively and are both used in~\cref{alg:ggns}. The cluster statistics formalism presented in this section follows~\citep{handley2015polychord}. We refer the reader to the original paper for derivations on where these formulas come from. 
}

{
The summary statistics are initiated using the following equation:
}

{
\begin{align}
    \label{eq:ini0}
    \overline{Z} &= 0, \\ 
    \label{eq:ini1}
    \overline{Z_p} &= \left\{ \overline{Z} \right\}, \\ 
    \label{eq:ini2}
     \overline{Z^2} &= 0, \\ 
     \label{eq:ini3}
    \overline{Z_p^2} &= \left\{ \overline{Z^2} \right\}, \\ 
    \label{eq:ini4}
     \overline{Z X} &= 0, \\
     \label{eq:ini5}
    \overline{Z X_p} &= \left\{ \overline{Z X} \right\}, \\ 
    \label{eq:ini6}
    \overline{Z_p X_p} &= \left\{ \overline{Z X} \right\}, \\ 
    \label{eq:ini7}
     \overline{X} &= 1, \\ 
     \label{eq:ini8}
    \overline{X_p} &= \left\{ \overline{X} \right\}, \\ 
    \label{eq:ini9}
    \overline{X_p^2} &= \left\{ \overline{X^2} \right\}, \\ 
    \label{eq:ini10}
     \overline{X_p X_q} &= 0 \qquad (q\ne p), \\
     \label{eq:ini11}
\end{align}
where $p$ and $q$ refer to the cluster numbers, initially $1$. To update the summary statistics, we use: 
}

{
\begin{align}
    \label{eq:update0}
  \overline\ev &\to \overline\ev + \frac{\overline X_p \lik}{n_p+1},\\
    \label{eq:update1}
  \overline\ev_p &\to \overline\ev_p + \frac{\overline X_p \lik}{n_p+1},\\
    \label{eq:update2}
  \overline X_p &\to \frac{n_p\overline X_p }{n_p+1}, \\
    \label{eq:update3}
  \overline{\ev^2} &\to \overline{\ev^2} + \frac{2\overline{\ev X_p}\lik_p}{n_p+1}  + \frac{2\overline{X_p^2}\lik^2}{(n_p+1)(n_p+2)}, \\
    \label{eq:update4}
  \overline{\ev_p^2} &\to \overline{\ev_p^2} + \frac{2\overline{\ev_p X_p}\lik}{n_p+1}  + \frac{2\overline{X_p^2}\lik^2}{(n_p+1)(n_p+2)}, \\
    \label{eq:update5}
  \overline{\ev X_p} &\to \frac{n_p\overline{\ev X_p}}{n_p+1}  + \frac{n_p\overline{X_p^2} \lik}{(n_p+1)(n_p+2)},   \\
    \label{eq:update6}
  \overline{\ev X_q} &\to \overline{\ev X_p}  + \frac{\overline{X_p X_q} \lik}{(n_p+1)} \qquad (q\ne p),  \\
    \label{eq:update7}
  \overline{\ev_p X_p} &\to \frac{n_p\overline{\ev_p X_p}}{n_p+1}  + \frac{n_p\overline{X_p^2} \lik}{(n_p+1)(n_p+2)},   \\
    \label{eq:update8}
  \overline{X_p^2} &\to \frac{n_p\overline{X_p^2}}{n_p+2}, \\
    \label{eq:update9}
  \overline{X_p X_q} &\to \frac{n_p\overline{X_p X_q}}{n_p+1} \qquad (q\ne p).
\end{align}
}

{
When we need to split a cluster $p$ into multiple clusters $i$, we use: 
}

{
\begin{align}
    \label{eq:new0}
  \overline{X_i}&=  \frac{n_i}{n} \overline X_p, \\
    \label{eq:new1}
  \overline{X_i^2}&= \frac{n_i(n_i+1)}{n(n+1)} \overline{X_p^2}, \\
    \label{eq:new2}
  \overline{X_i X_j} &= \frac{n_i n_j}{n(n+1)} \overline{X_p^2}, \\
    \label{eq:new3}
  \overline{X_i Y} &= \frac{n_i}{n} \overline{X_p Y} \qquad Y\in \{Z,Z_p,X_q\}, \\ 
    \label{eq:new4}
  \overline{Z_i}&=  \frac{n_i}{n} \overline Z_p, \\
    \label{eq:new5}
  \overline{Z_i X_i}&= \frac{n_i(n_i+1)}{n(n+1)} \overline{Z_p X_p}, \\
    \label{eq:new6}
  \overline{Z_i^2}  &= \frac{n_i(n_i+1)}{n(n+1)} \overline{Z_p^2}. \\
\end{align}
}

{
\section{Termination Criterion Explained} \label{app:termination}
}

\begin{figure}[t]
\begin{minipage}{0.49\textwidth}
    \caption{In nested sampling, the likelihood $\lik$ (dotted line) goes up, as the algorithm progresses from $X=1$ to $X=0$ (right to left in the plot). However, the product $X \lik (X)$ starts low, as the likelihood is small, peaks, end goes down again. Therefore, our termination criterion checks when $X \lik (X)$ has peaked, and again gone close to zero. Image credit~\citep{handley2015polychord}.}
\end{minipage}\hfill
\raisebox{-1em}{
\begin{minipage}{0.49\textwidth}
\includegraphics[width=\linewidth]{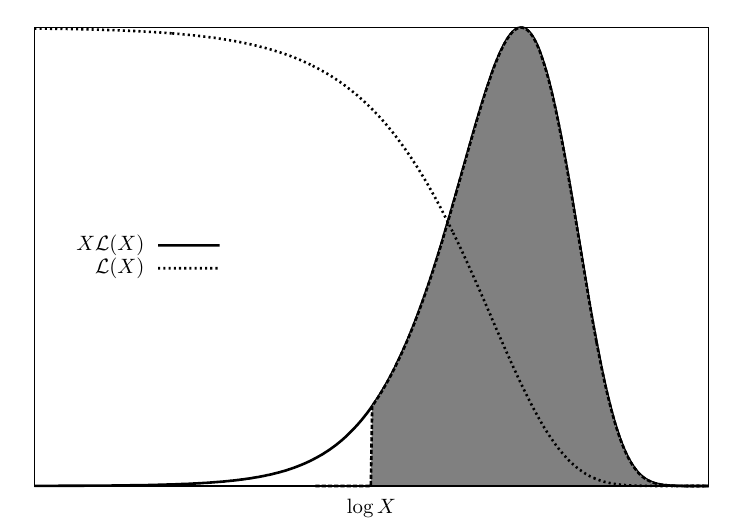}
\end{minipage}
}
    \label{fig:termination}
 \end{figure}

{
\cref{fig:termination} explains the intuition behind the new termination criterion introduced in~\cref{sec:implementation}. In nested sampling, the likelihood increases as the algorithm progresses, corresponding to going from right ($X=1$) to left ($X=0$) in the figure. However, the product of the prior volume $X$ and the likelihood $\lik$ always follow the same pattern: It starts low, reaches a peak, and then decreases towards zero again. Therefore, our termination criterion consists of checking the ratio of the current value of $X \lik$ to the maximum value that $X \lik$ has reached throughout the algorithm. When that fraction is smaller than some threshold, we stop the algorithm. 
}

{
\section{Evidence Estimation as a Function of Likelihood Evaluations}
}

{
\begin{figure}[t]
    \centering
    \includegraphics[width=0.49\linewidth]{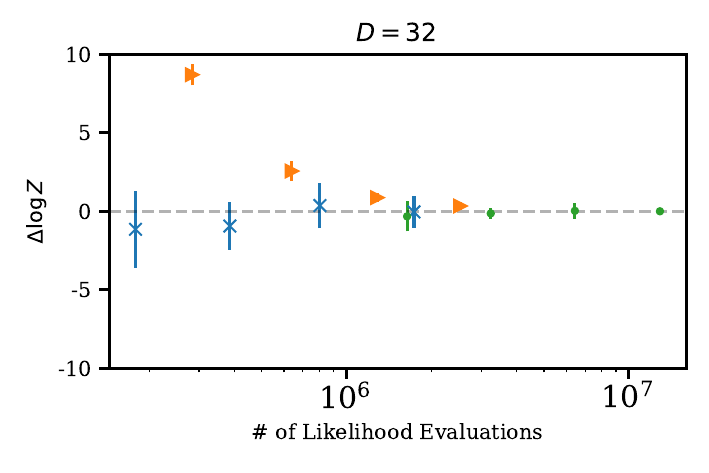}
    \includegraphics[width=0.49\linewidth]{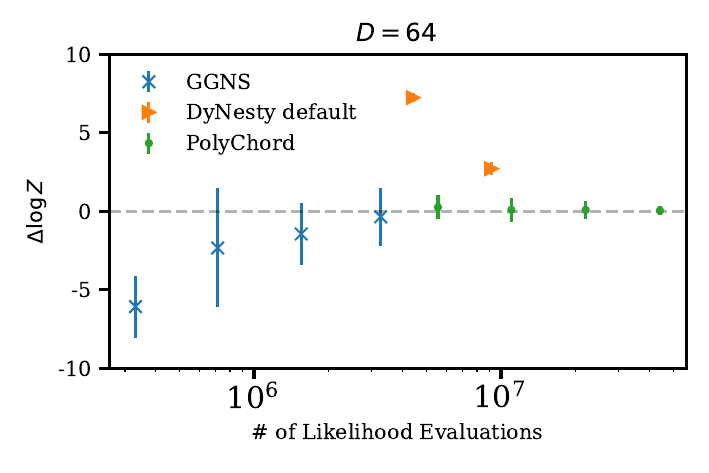}
    \caption{The bias in the estimate of $\log Z$ as a function of number of likelihood evaluations, for \algoname (blue), and other nested sampling algorithms ({\tt PolyChord} in green and {\tt dynesty} in orange}). We achieve different numbers of like evaluations by changing the number of live points. We run each algorithm with $n_{\rm live} = \left\{ 50, 100, 200, 400 \right\}$. Note that the {\tt dynesty} runs with $n_{\rm live} = 50 $ and $n_{\rm live} = 100$ are not in the plot, as they are too far up in the y-axis. 
\end{figure}
}

{
In this appendix, we study the relationship between the number of likelihood evaluations, and the estimate of $\Delta \log Z$. The easiest way to vary the number of likelihood evaluations, is by varying the number of live points used. We repeat the analysis of~\cref{sec:comparison}, for two values of the number of dimensions $d = 32$ and $d = 64$, for each of the algorithms; varying the number of live points in the range
$n_{\rm live} = \left\{ 50, 100, 200, 400 \right\}$. 
}

{
We see how, in both cases, {\tt dynesty} can lead to very biased inference, if the number of live points is low. On the other hand, {\tt PolyChord} reliably achieves unbiased inference, at the expense of a much higher number of likelihood evaluations. \algoname gets the best of each, by achieving unbiased inference with less likelihood evaluations. We also see how our algorithm scales better with dimensionality, when we compare its performance between the left and the right plots, with the other algorithms. 
}
\end{document}

%% file: math_commands.tex
\usepackage{amsmath,amsfonts,bm}

\def\eqref#1{equation~\ref{#1}}

\def\1{\bm{1}}

\DeclareMathAlphabet{\mathsfit}{\encodingdefault}{\sfdefault}{m}{sl}
\SetMathAlphabet{\mathsfit}{bold}{\encodingdefault}{\sfdefault}{bx}{n}

\def\gD{{\mathcal{D}}}

\def\gL{{\mathcal{L}}}

